\documentclass[conference]{IEEEtran}
\IEEEoverridecommandlockouts
\usepackage{cite}
\usepackage{amsmath,amssymb,amsfonts}
\usepackage{algorithmic}
\usepackage{graphicx}
\usepackage{textcomp}
\usepackage{xcolor}

\usepackage{tikz}
\usetikzlibrary{positioning}
\usetikzlibrary{shapes.geometric, arrows}
\usetikzlibrary{calc}
\usetikzlibrary{fit}
\tikzstyle{startstop} = [rectangle, rounded corners, minimum width=3cm, minimum height=1cm,text centered, draw=black, fill=red!30]
\tikzstyle{process} = [rectangle, minimum width=3cm, minimum height=1cm, text centered, draw=black, fill=orange!30]
\tikzstyle{decision} = [diamond, minimum width=3cm, minimum height=1cm, text centered, draw=black, fill=green!30]

\usepackage{graphicx}

\usepackage{subfigure}    
\usepackage{float}
\usepackage{svg}
\usepackage{url}
\usepackage{threeparttable}

\usepackage{siunitx}
\usepackage{hyperref}
\definecolor{linkblue}{RGB}{0, 102, 204}
\hypersetup{
    colorlinks=true,
    urlcolor=linkblue
}
\usepackage{cleveref}
\usepackage{glossaries}
\usepackage{comment}
\usepackage{adjustbox}

\newglossaryentry{ICP}
{
        name=ICP,
        description={Iterative closest point.}
}

\newacronym{cpd}{CPD}{Coherent Point Drift}

\def\BibTeX{{\rm B\kern-.05em{\sc i\kern-.025em b}\kern-.08em
    T\kern-.1667em\lower.7ex\hbox{E}\kern-.125emX}}
\begin{document}

\title{Comparing Commercial Depth Sensor\\Accuracy for Medical Applications
}

\author{
Pit Henrich$^{1}$,
Maximilian Weiherer$^{2}$,
Franziska Hansen$^{1}$,
Bernhard Egger$^{2}$,
Franziska Mathis-Ullrich$^{1*}$%
\\[1ex]
\small
$^{1}$Department of Artificial Intelligence in Biomedical Engineering,
Friedrich-Alexander-Universit\"at Erlangen-N\"urnberg, Erlangen, Germany\\
$^{2}$Department of Computer Science,
Friedrich-Alexander-Universit\"at Erlangen-N\"urnberg, Erlangen, Germany\\
$^{*}$Corresponding author: franziska.mathis-ullrich@fau.de
}

\maketitle

\begin{abstract}
    Depth estimation has numerous medical and surgical applications.
    We benchmark four depth sensors on a porcine bone specimen, a porcine belly specimen, and a silicone kidney phantom using stylus-sampled references.
    These objects contain several real-world challenges, including homogeneous surfaces, specular surfaces, and subsurface scattering.
    The comparison includes stereo, structured-light, and time-of-flight sensors at a distance of approximately \SI{50}{\centi\metre}.
    Specifically, the Intel RealSense D405 (Intel RealSense, United States), PMD Flexx2 (pmdtechnologies, Germany), Stereolabs ZED 2i (Stereolabs, France), and Zivid 2M+ 60 (Zivid, Norway) are compared.
    The Zivid 2M+ 60 performed best across all objects and metrics considered in this work. The ZED ranked second for real tissue, but last on the phantom.
    \textbf{Data:} \url{https://somefoo.github.io/schweinebauch}
\end{abstract}


\section{Introduction}
    
        \begin{figure}[tb]
        \centering
        \begin{tikzpicture}
            \node[inner sep=0pt, rounded corners=15pt, clip] (img) {%
                \includegraphics[width=\columnwidth]{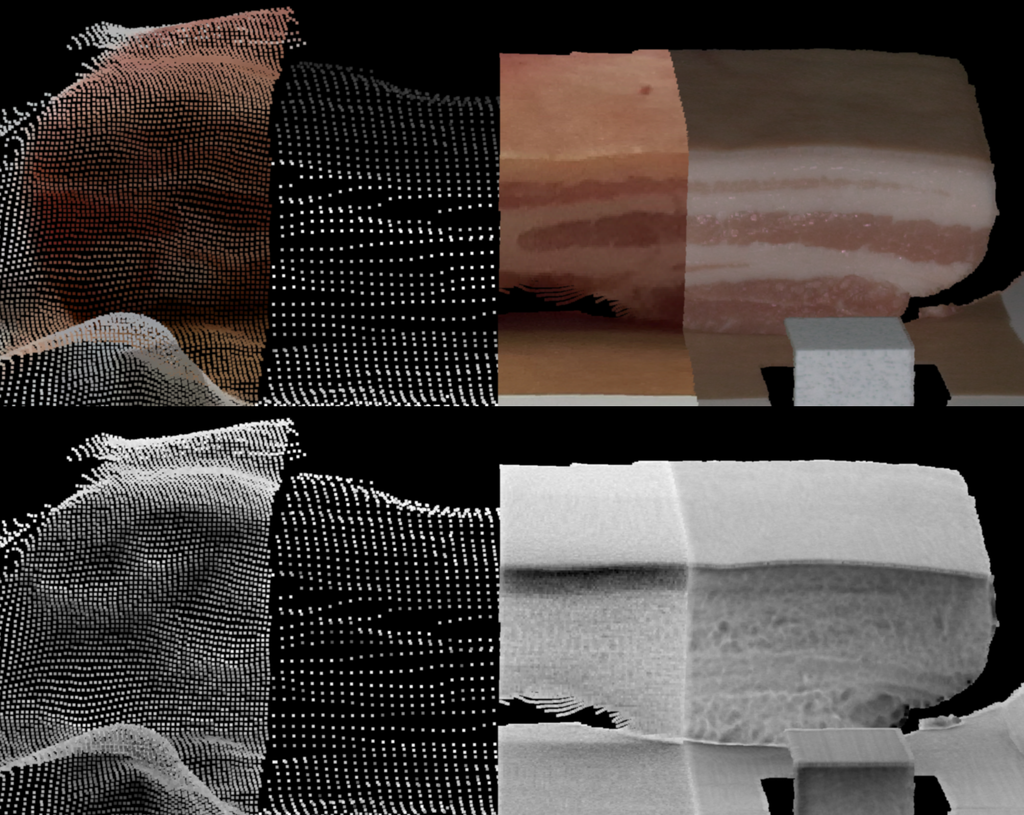}%
            };
    
            \node[anchor=center, text=white] at ([xshift=0, yshift=-2mm]img.north) {\small Color};
            \node[anchor=center, text=white] at ([xshift=0, yshift=-2mm]img.center) {\small Structure Only};
    
            \node[below=1mm of img.south west, anchor=north, xshift=0.15\linewidth] {D405};
            \node[below=1mm of img.south west, anchor=north, xshift=0.37\linewidth] {Flexx};
            \node[below=1mm of img.south west, anchor=north, xshift=0.58\linewidth] {ZED};
            \node[below=1mm of img.south west, anchor=north, xshift=0.83\linewidth] {Zivid};
        \end{tikzpicture}
        \caption{Examples of point clouds captured by the evaluated commercial depth sensors.
        The top shows the colored point clouds produced by each sensor. The Flexx only provides grayscale values. The bottom shows the same point clouds without color.
        }
        \label{fig:qualitative-point-clouds}
    \end{figure}
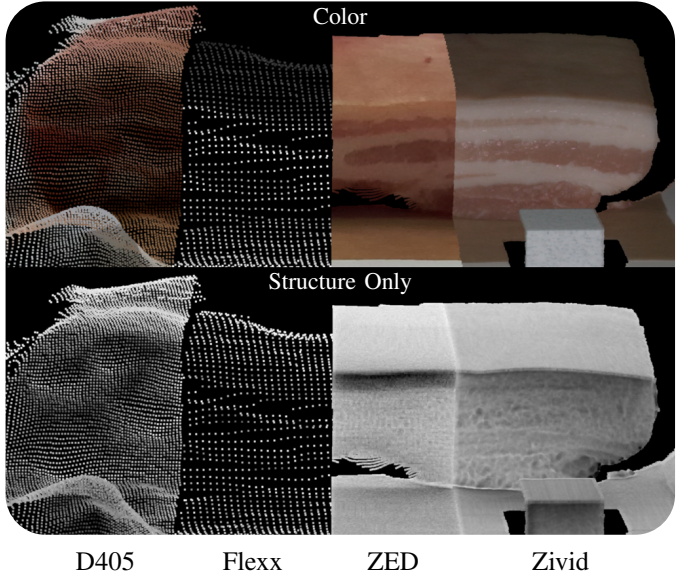

    \noindent 3-D surface information is central for automating many medical workflows, enabling tasks such as anatomy localization~\cite{henrich2024looc}, surgical outcome assessment~\cite{max1}, tissue tracking~\cite{henrich2025ludo}, and robotic assistance~\cite{gyenes2025}.
    Burger et al.~\cite{burger2023comparative} compare different commercial depth sensors for close-range surgical simulation.
    However, their work only focuses on stereo camera systems and does not provide a comparison to a geometric representation of real tissue.
    Villa et al.~\cite{villa2025benchmarking} assess multiple commercial depth sensors for neurosurgical applications.
    Their evaluation is conducted on a synthetic phantom that, while geometrically accurate, has a substantial visual difference to real tissue.
    Further work compares depth sensors~\cite{servi2021metrological, curto2022experimental}; however, none of them consider a medical application.
    This work evaluates four commercial sensors on real porcine tissue: the Intel RealSense D405 (Intel RealSense, United States), Stereolabs ZED 2i - \SI{2.1}{\milli\metre} (Stereolabs, France), Zivid 2M+ 60 (Zivid, Norway), and PMD Flexx2 (pmdtechnologies, Germany), against stylus-sampled reference point clouds.
    The contribution of this paper is a quantitative comparison across sensor principles and object types, example sensor outputs are shown in \Cref{fig:qualitative-point-clouds}.

\section{Setup and Experiments}

        \begin{figure}[t]
        \centering
        \includegraphics[width=1.0\columnwidth]{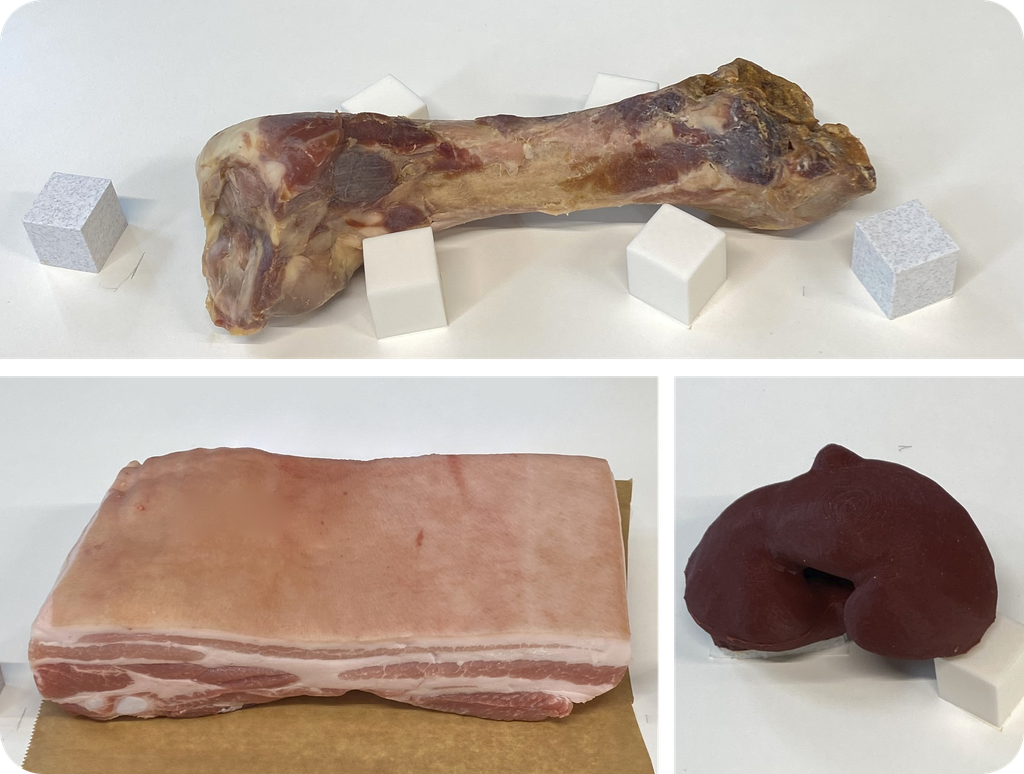}
        \caption{Evaluated objects: porcine bone specimen, porcine belly specimen, and silicone phantom. Widths (left to right in images) are approximately \SI{250}{\milli\metre}, \SI{230}{\milli\metre}, 
        and \SI{115}{\milli\metre}, respectively.
        }
        \label{fig:evaluated-objects}
    \end{figure}

        \begin{table*}[tb]
        \centering
        \caption{An overview of the depth sensors, their specifications, and the used settings and software versions.}
        \label{tab:hardware-settings-outline}
        \begin{adjustbox}{max width=\linewidth}
            \begin{tabular}{lllll}
            \hline
             & \textbf{RealSense D405} & \textbf{PMD Flexx2} & \textbf{Stereolabs ZED 2i (2.1mm)} & \textbf{Zivid 2M+ 60} \\
            \hline
            Principle 
            & Stereo 
            & Time-of-flight
            & Stereo / neural depth 
            & Structured light \\
            
            Oper. range 
            & 0.07--0.50\,m 
            & 0.10--4.0\,m
            & 0.30--12.0\,m 
            & 0.30--1.10\,m \\
 
            Megapixels
            & 0.92
            & 0.039
            & 2.74
            & 5.01 \\           

            Settings 
            & 848$\times$480, Default
            & Mode\_9\_5\_FPS, Auto Exposure
            & Neural Plus, HD2K
            & Inspection: SmallFeatures\\
            
            Firmware
            & 5.16.0.1
            & 6.9.0.3733
            & 1523
            & 1.31.3 \\
            
            SDK
            & 2.57.7.10387
            & 5.7.0.2331
            & 5.3.0
            & 2.14.2 \\
            
            \hline
            \end{tabular}
        \end{adjustbox}

        \begin{tikzpicture}
            \node[
                inner sep=0pt,
                rounded corners=8pt,
                clip
            ] (img) {
                \includegraphics[width=0.85\linewidth]{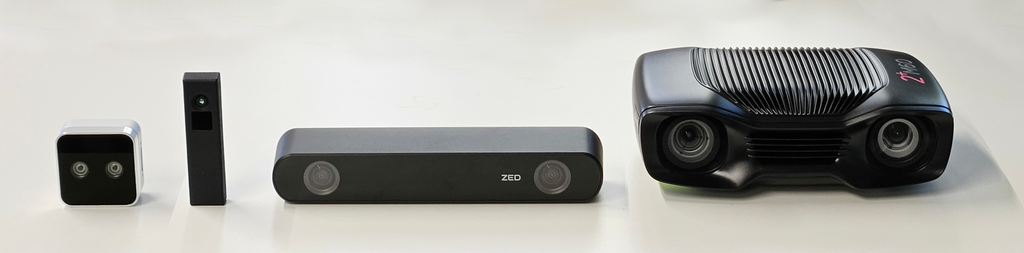}
            };

            \draw[->,black,dotted, thick]
                ($(img.north west)!0.18!(img.north east)+(0,0.1)$)
                -- ++(0,-0.25)
                -| ($(img.center)+(-6.2,0.3)$);
            
            \draw[->,black,dotted, thick]
                ($(img.north west)!0.33!(img.north east)+(0,0.1)$)
                -- ++(0,-0.45)
                -| ($(img.center)+(-4.65,1.1)$);
            
            \draw[->,black,dotted, thick]
                ($(img.north west)!0.573!(img.north east)+(0,0.1)$)
                -- ++(0,-0.25)
                -| ($(img.center)+(-1.1,0.2)$);
            
            \draw[->,black,dotted, thick]
                ($(img.north west)!0.82!(img.north east)+(0,0.1)$)
                -- ++(0,-0.25)
                -| ($(img.center)+(4.0,1.4)$);

        \end{tikzpicture}
    
    \end{table*}

    We compare four depth sensors on three objects: a porcine bone specimen with dried muscular tissue attached, a porcine belly specimen with heterogeneous soft-tissue layers, and a silicone phantom (see \Cref{fig:evaluated-objects}).
    Four 3-D-printed patterned marble-white PLA (Bambu Lab, China) calibration cubes have been placed around each object to enable registration between the stylus reference and depth-sensor captures.
    A pattern provides visual features for stereo sensors.
    Fully white cubes are used to fixate objects.
    Both cube types were fixed to the table using adhesive tape throughout stylus sampling and depth-sensor acquisition.
    The evaluated sensors and their acquisition settings are summarized in \Cref{tab:hardware-settings-outline}.
    The D405 is a close-range stereo sensor, the ZED 2i uses stereo with neural depth estimation, the Zivid structured light, and the Flexx2 infrared time-of-flight.
    The sensors are operated at an object distance of approximately \SI{50}{\centi\metre}.
    An additional D405 porcine belly close-up of approximately \SI{33}{\centi\metre} is performed as \SI{50}{\centi\metre} is at the upper limit of the D405.
    A distance of \SI{33}{\centi\metre} ensures that the calibration cubes are still visible.

        \begin{figure}[t]
        \centering
        \resizebox{1.0\columnwidth}{!}{%
        \begin{tikzpicture}[
            process/.style={
                rounded corners,
                align=center,
                inner sep=4pt
            },
            arrow/.style={
                ->,
                thick,
                rounded corners=8pt
            }
        ]
        \newcommand{\roundedimage}[2][2.4cm]{%
            \begin{tikzpicture}
                \node[
                    anchor=south west,
                    inner sep=0,
                    rounded corners=8pt,
                    clip
                ] at (0,0) {\includegraphics[width=#1]{#2}};
            \end{tikzpicture}
        }
        \newcommand{\imgnode}[4]{%
            \node[process] (#1) at (#2) {%
                \roundedimage{#3}\\[0mm]
                #4%
            };
        }
    
        \imgnode{stylus}{0,1.7}
            {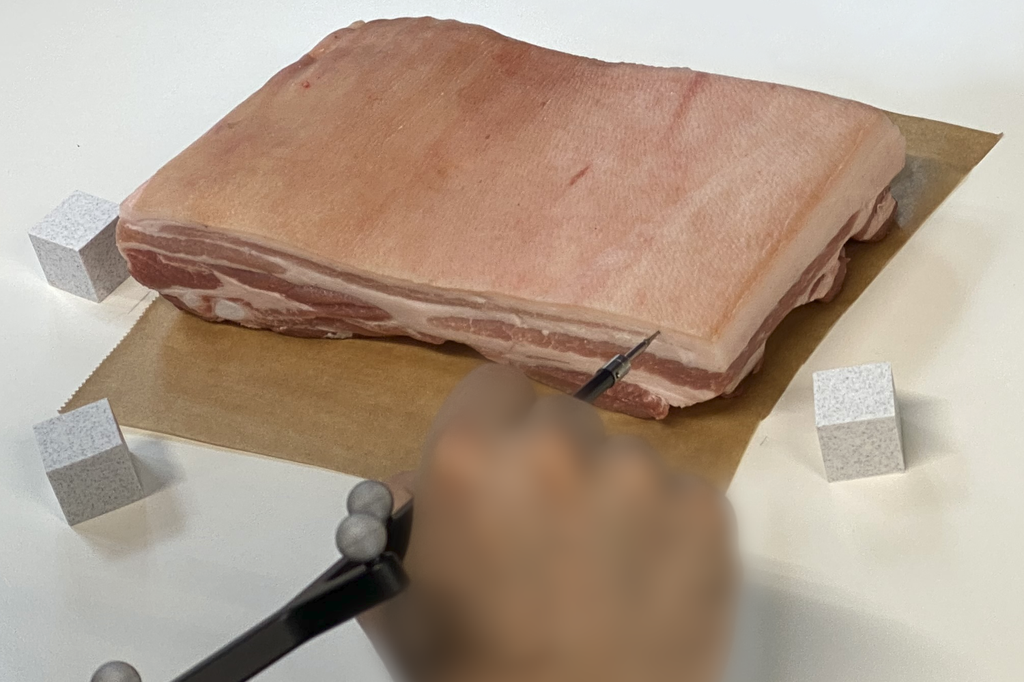}
            {Stylus Sampling}
    
        \imgnode{depth}{4.2,1.7}
            {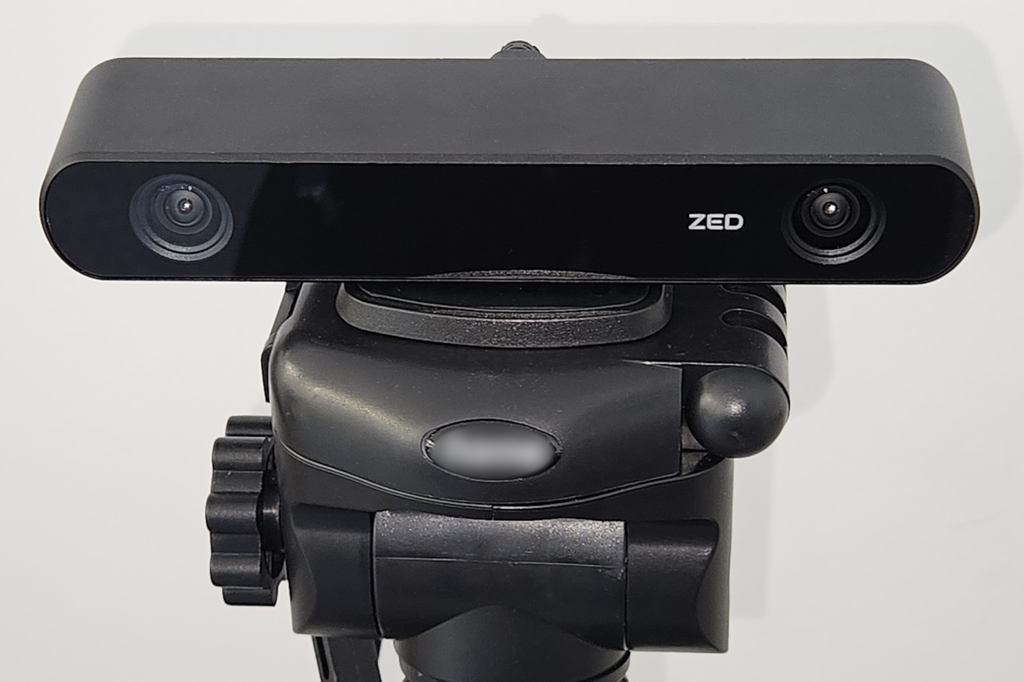}
            {Depth Sensor}
    
        \imgnode{output}{0,-1.2}
            {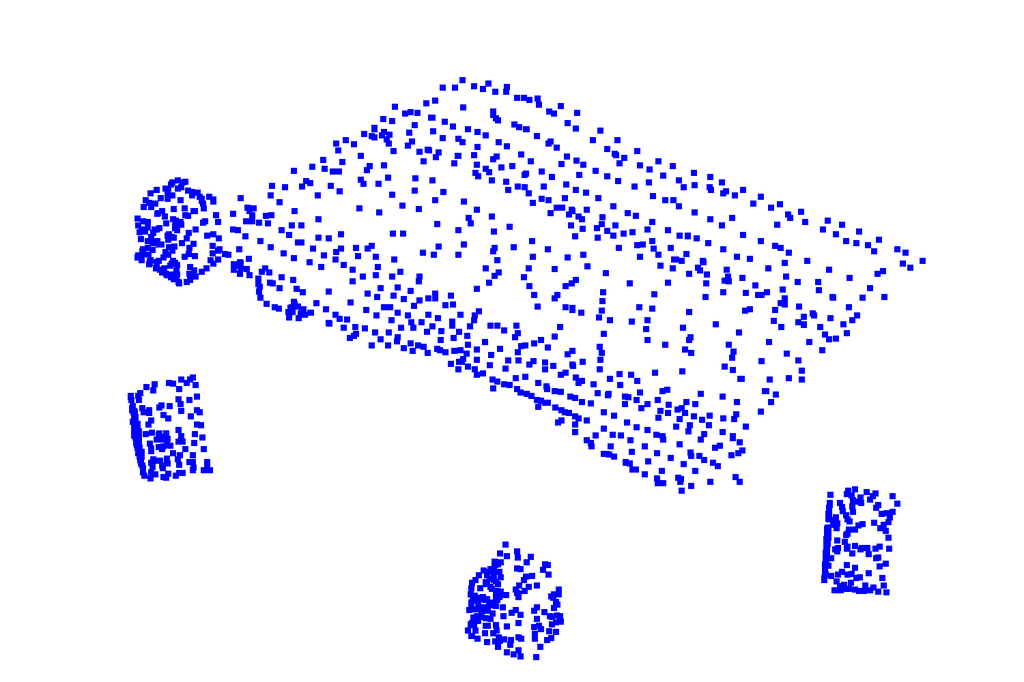}
            {Sampled Point Cloud}
    
        \imgnode{registered}{4.2,-1.2}
            {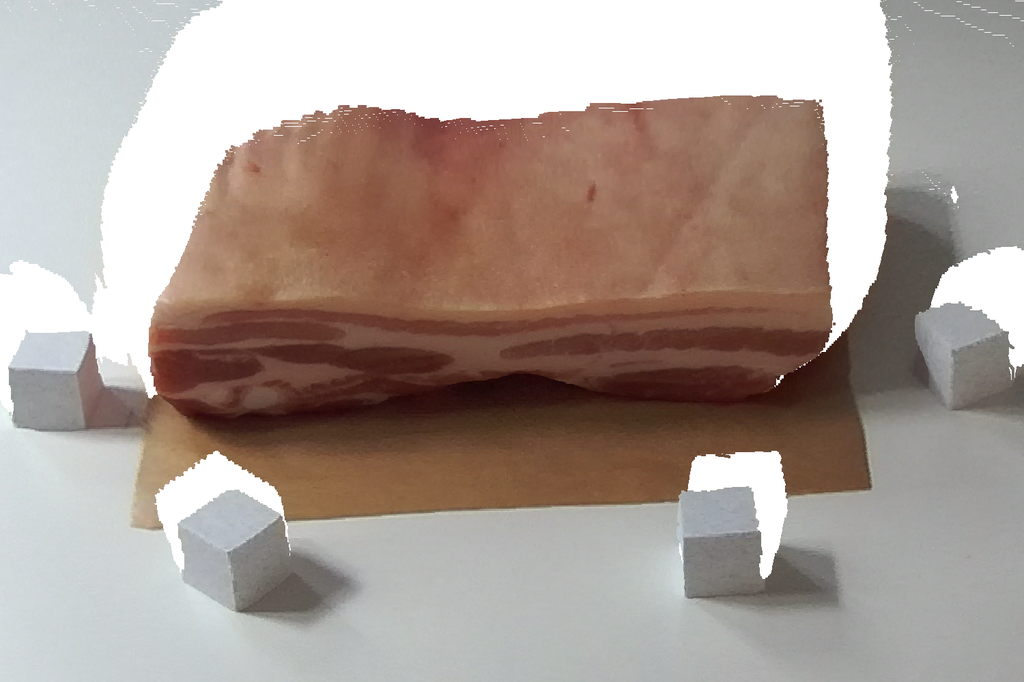}
            {Captured Point Cloud}
    
        \imgnode{merged}{2.1,-4.1}
            {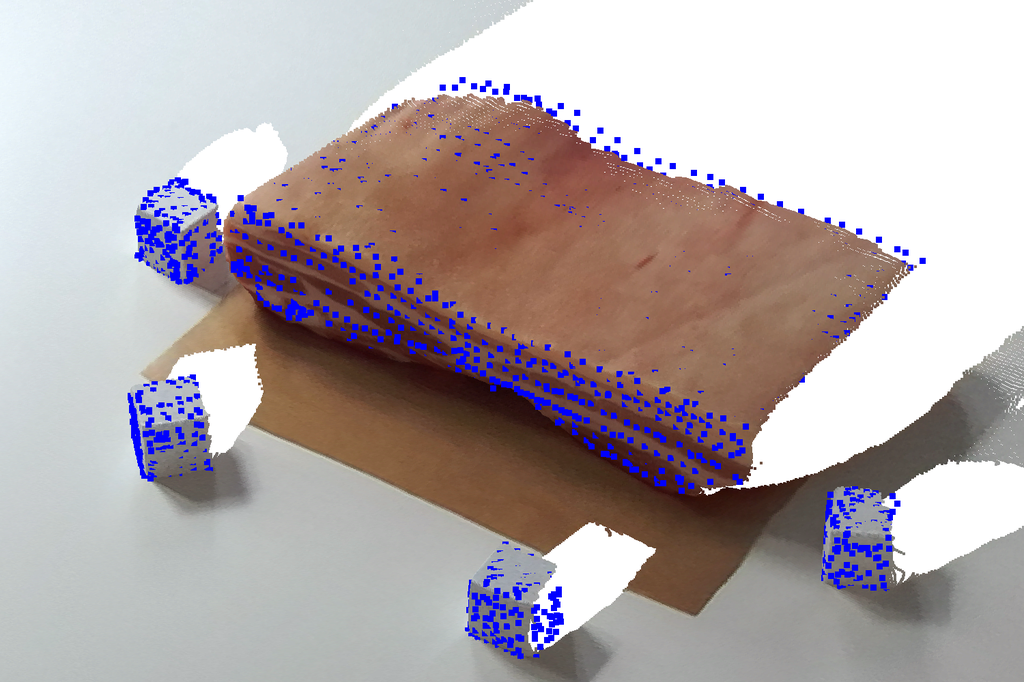}
            {Registered Point Clouds}
    
        \draw[arrow] (stylus.south) -- (output.north);
        \draw[arrow] (depth.south) -- (registered.north);
        \draw[arrow]
            (output.south) |-
            node[pos=0.2, sloped, above] {\small \textcolor{gray}{Target}}
            (merged.west);
        \draw[arrow]
            (registered.south) |-
            node[pos=0.2, sloped, above] {\small \textcolor{gray}{Source}}
            (merged.east);
        \end{tikzpicture}
        }
        \caption{Stylus reference acquisition, depth capture, and point cloud registration. The registration is performed using only the calibration cubes.}
        \label{fig:data-collection-workflow}
    \end{figure}

    As shown in \Cref{fig:data-collection-workflow}, reference geometry is acquired with a stylus tracked by three OptiTrack PrimeX 13W cameras (NaturalPoint, United States) at a distance of approximately \SI{1.5}{\metre}.
    The tracker and stylus are re-calibrated for each object.
    Across the three objects, the OptiTrack self-reported stylus error was \SI{0.59 \pm 0.10}{\milli\metre} \textit{(mean $\pm$ standard deviation)}. %
    The manual stylus sampling targeted the calibration cubes and parts of the object that we expected to be visible to the sensors, and produced point clouds containing $446$, $649$, and $1412$ total points for the phantom, bone, and porcine belly scenes, of which $134$, $281$, and $798$ belong to the actual object.
    Each sensor acquires three point clouds per object to capture inter-acquisition variability, and each point cloud is first manually aligned and then registered to the reference using rigid coherent point drift.
    For alignment and registration, only the calibration cubes are used.
    We then compute per-point errors for each point in the reference point cloud, defined as the Euclidean distance between the reference point and the closest point on the registered point cloud.
    Based on these distances and for every capture, the first quartile $Q_1$, third quartile $Q_3$, and interquartile range $\mathrm{IQR}=Q_3-Q_1$ are calculated and then used to prune away reference points with a distance greater than the Tukey upper fence, $Q_3 + 1.5\cdot\mathrm{IQR}$, to effectively remove points that are not seen by the depth sensor.
    Accuracy is assessed based on per-point errors computed from the pruned reference geometry and registered captures.
    Finally, the Flexx has a lower resolution compared to the other sensors.
    We subsample sensor point clouds to one point per \SI{3}{\milli\metre^3} before computing errors to improve fairness.
    This ensures that all point clouds approximately match the density of point clouds acquired by Flexx, which exhibit a point spacing between \SI{2}{\milli\metre} and \SI{3}{\milli\metre} at a distance of \SI{50}{\centi\metre}.

    \colorlet{myorange}{green!55!white}

\newcommand{\roundedimage}[2][2.4cm]{%
    \begin{tikzpicture}
        \node[
            inner sep=0pt,
            rounded corners=6pt,
            clip
        ] (img) {\includegraphics[width=#1]{#2}};

        \draw[-, color=myorange] ($(img.south west)+(0.2cm,0.2cm)$)  -- ($(img.center)+(0.13cm,0.57cm)$);

        \filldraw[myorange] ($(img.center)+(0.13cm,0.57cm)$) circle (1.3pt);

        \node[
            anchor=south west,
            inner sep=0pt,
            rounded corners=6pt,
            clip,
        ] (zoom) at (img.south west) {%
            \includegraphics[
                width=1.5cm,
                trim=22cm 28cm 20cm 11cm,
                clip,
            ]{#2}%
        };

    \node[draw=myorange, thick, fit=(zoom), rounded corners=6pt]{};
        
    \end{tikzpicture}%
}

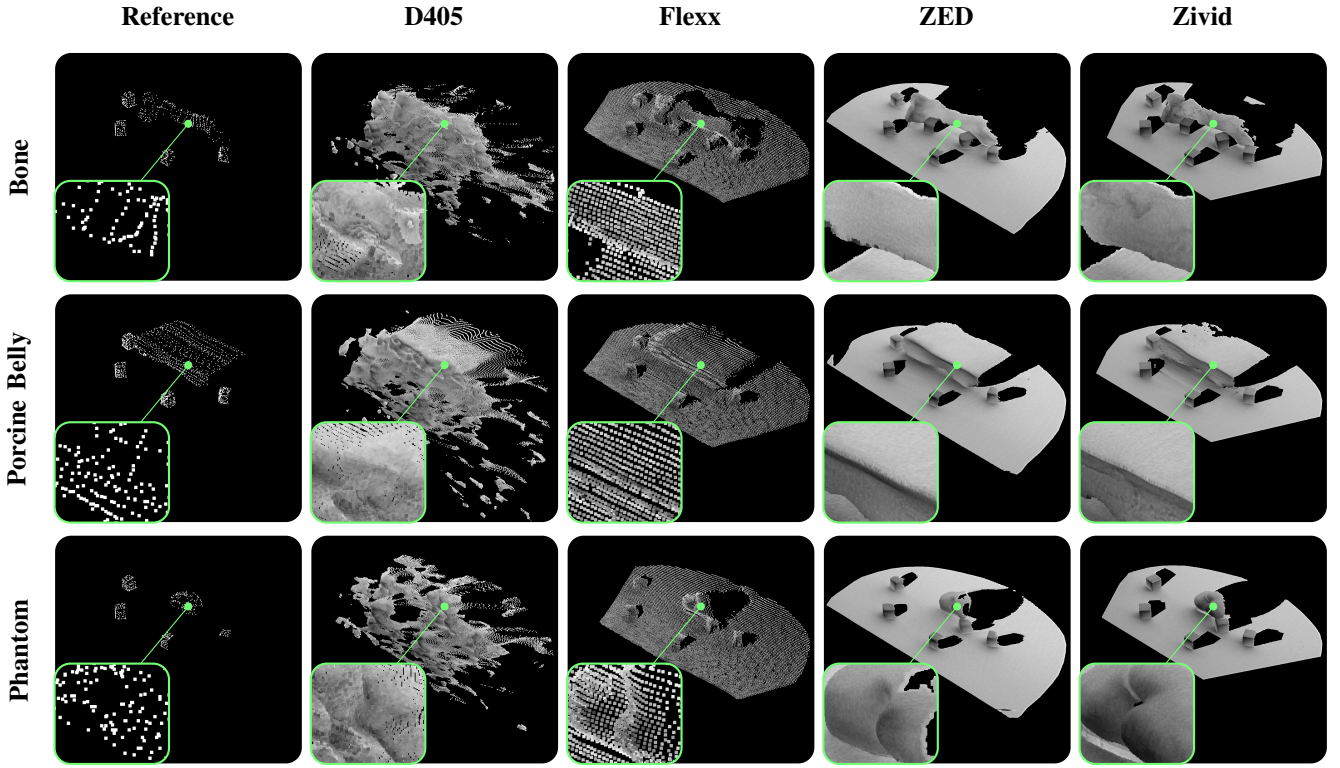
\begin{figure*}[ht]
    \centering

    \begin{tikzpicture}[
        every node/.style={inner sep=0pt}
    ]

        \def\imgw{0.18\textwidth}
        \def\xgap{0.005\textwidth}
        \def\ygap{0.008\textwidth}

        \node (b1)
            {\roundedimage[\imgw]
            {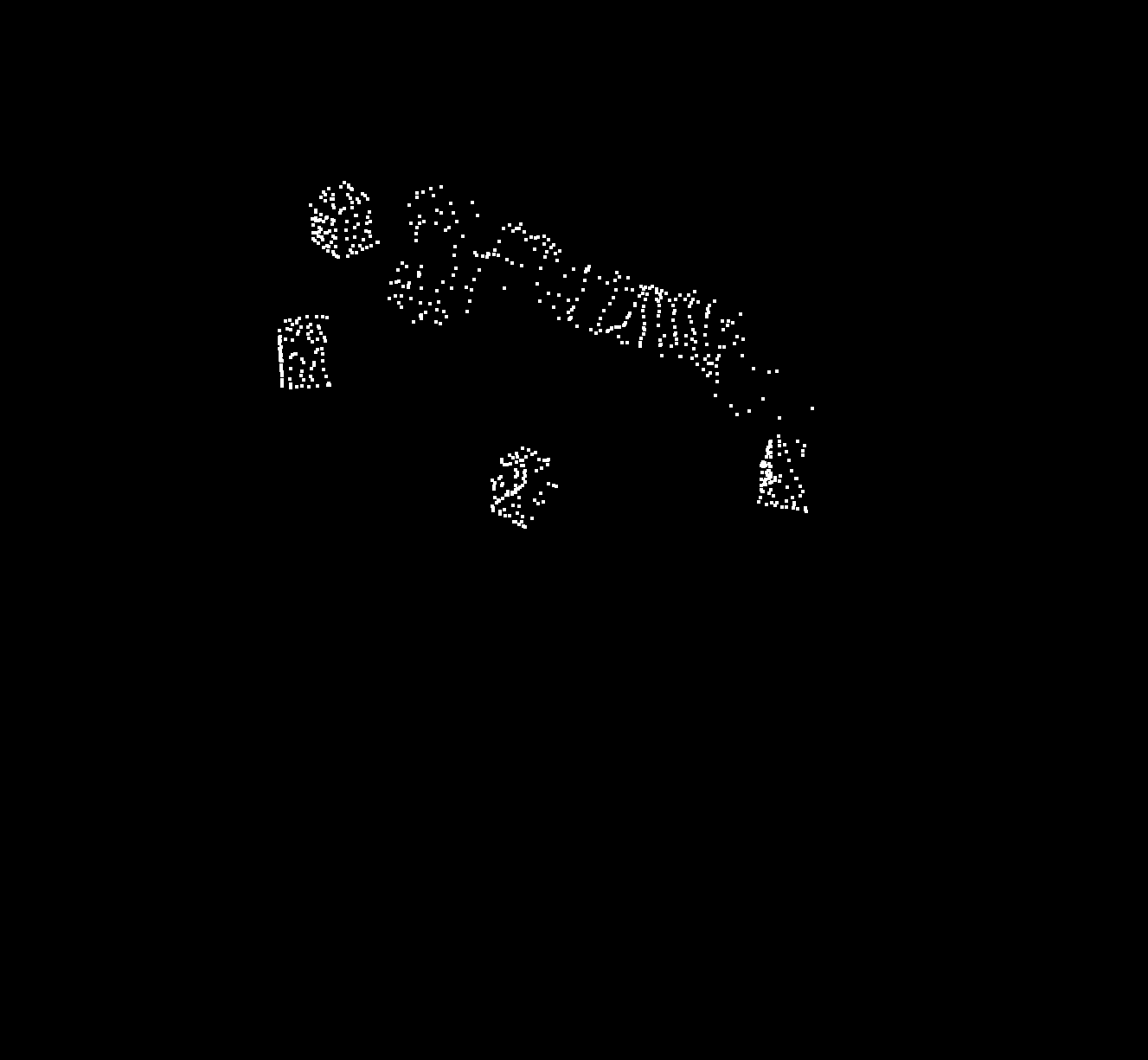}};

        \node[right=\xgap of b1] (b2)
            {\roundedimage[\imgw]
            {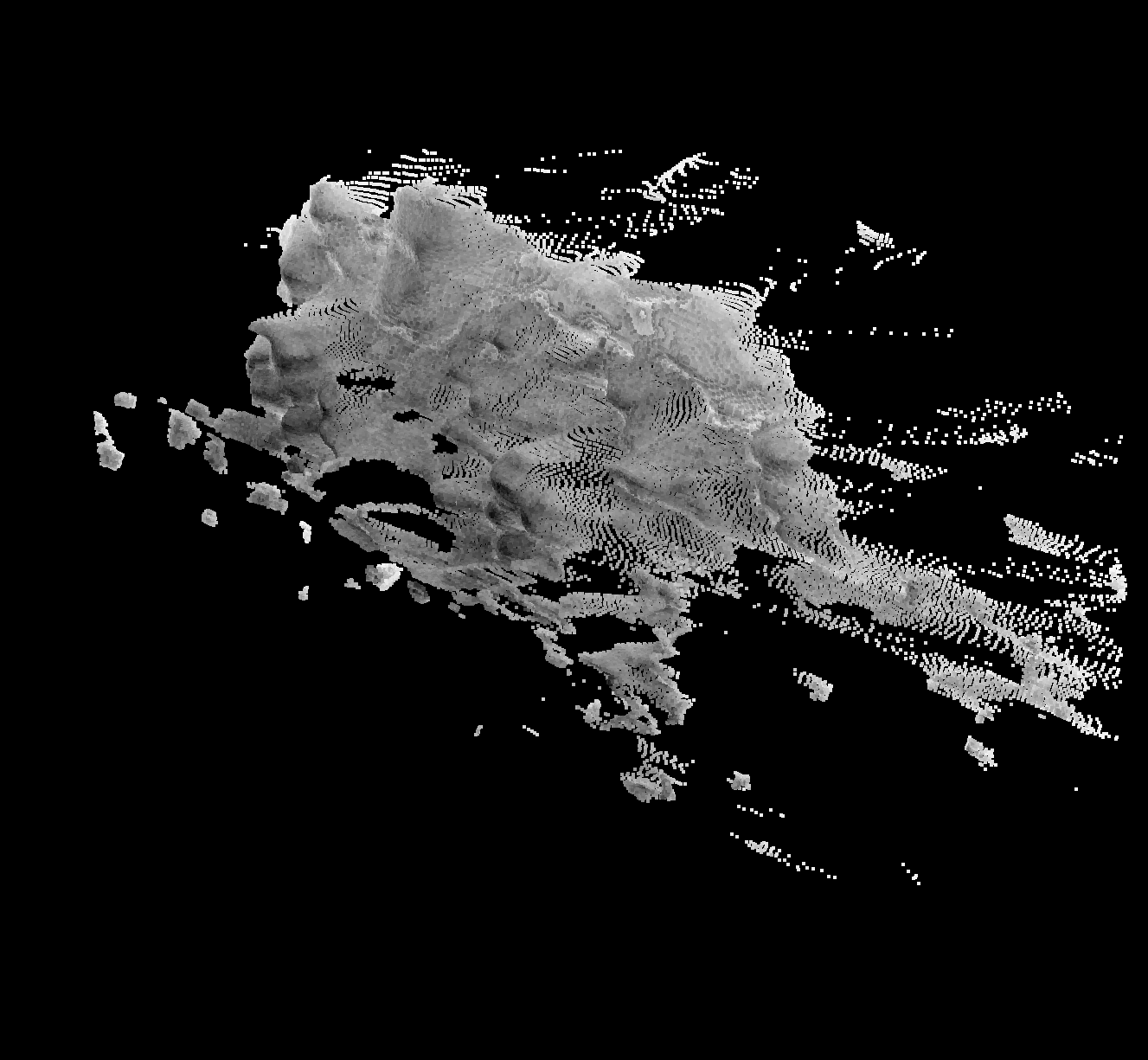}};

        \node[right=\xgap of b2] (b3)
            {\roundedimage[\imgw]
            {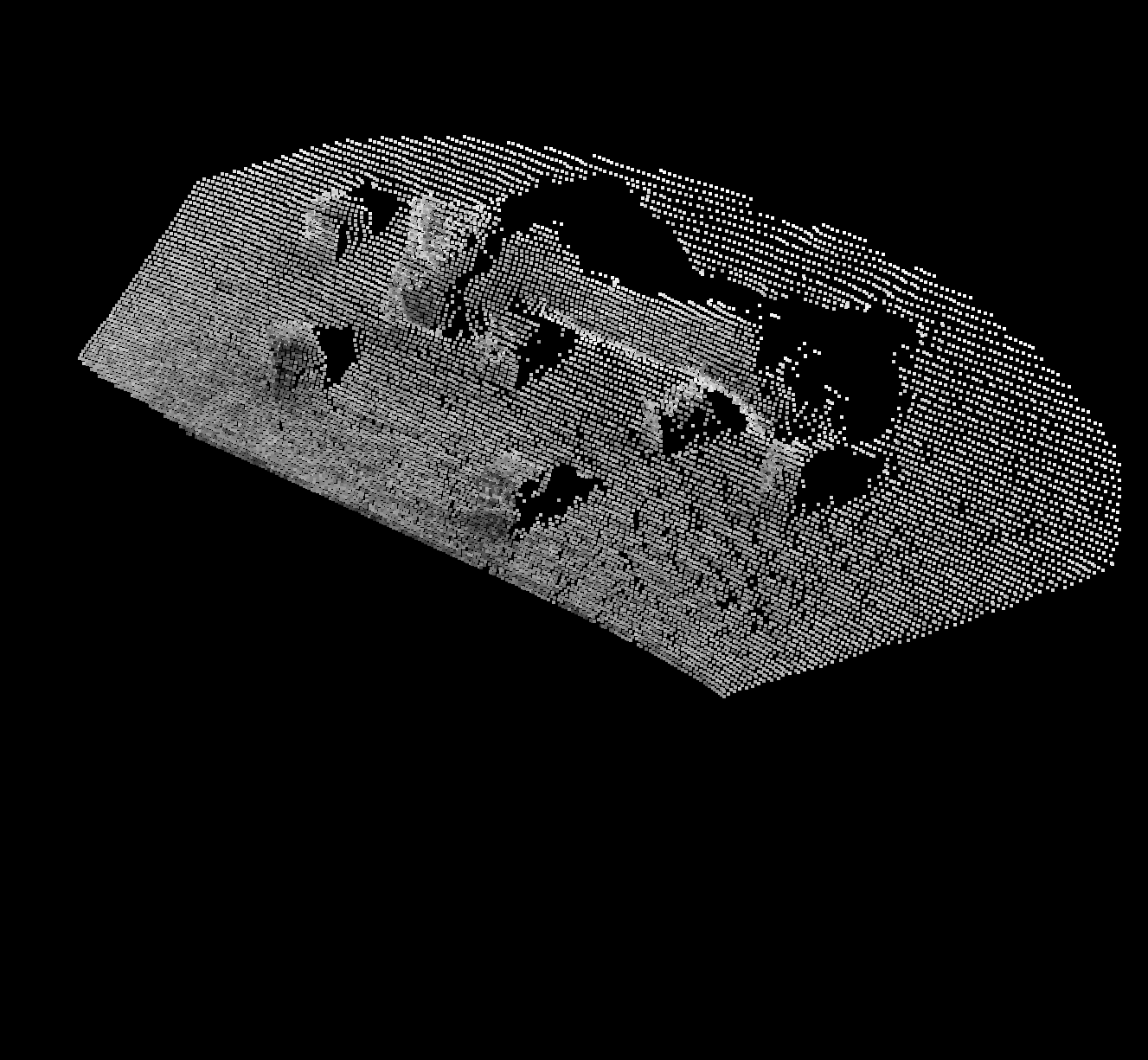}};

        \node[right=\xgap of b3] (b4)
            {\roundedimage[\imgw]
            {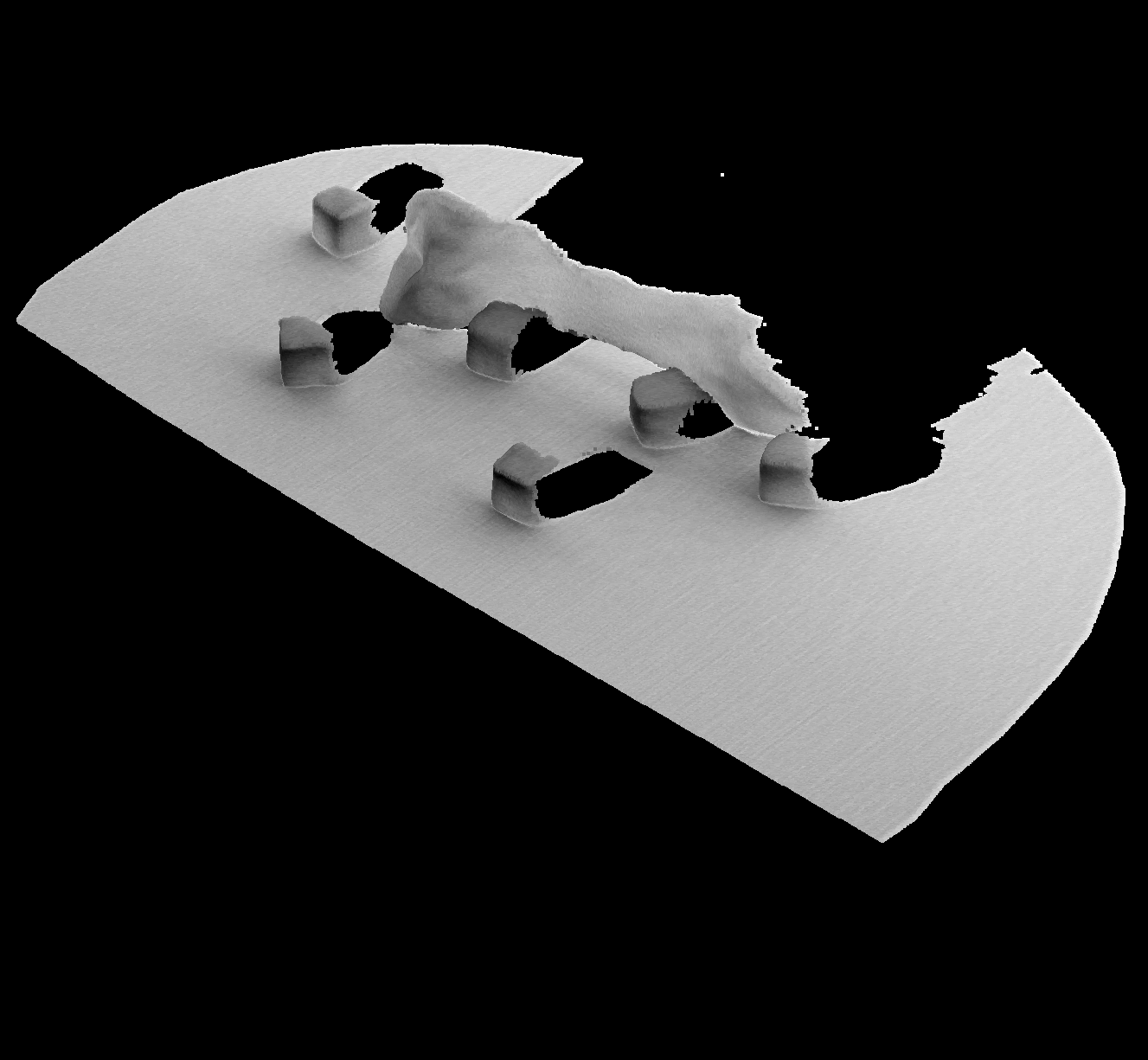}};

        \node[right=\xgap of b4] (b5)
            {\roundedimage[\imgw]
            {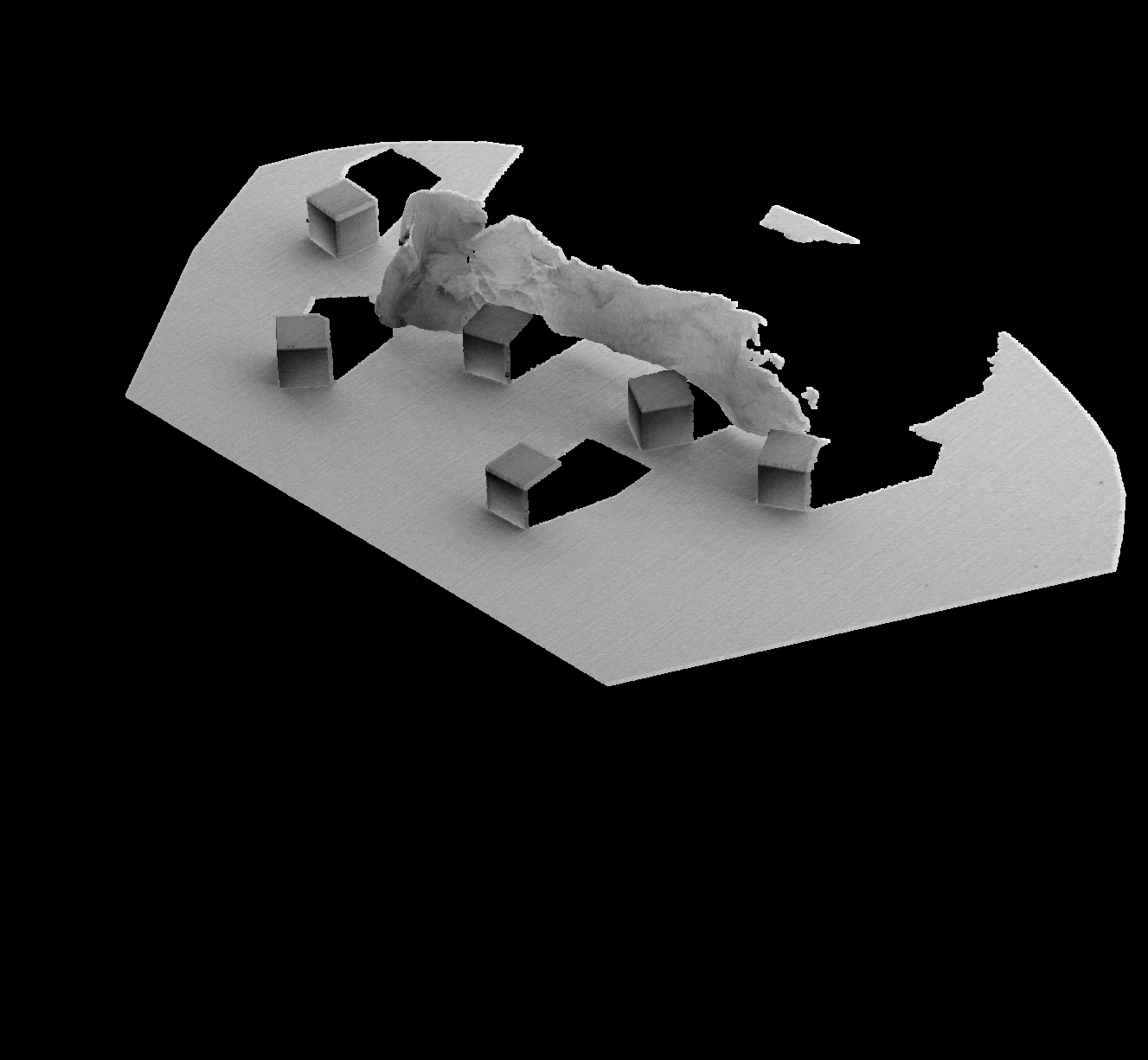}};

        \node[font=\bfseries, above=0.35cm of b1] {Reference};
        \node[font=\bfseries, above=0.35cm of b2] {D405};
        \node[font=\bfseries, above=0.35cm of b3] {Flexx};
        \node[font=\bfseries, above=0.35cm of b4] {ZED};
        \node[font=\bfseries, above=0.35cm of b5] {Zivid};

        \node[font=\bfseries, rotate=90]
            at ([xshift=-0.45cm]b1.west) {Bone};

        \node[below=\ygap of b1] (m1)
            {\roundedimage[\imgw]
            {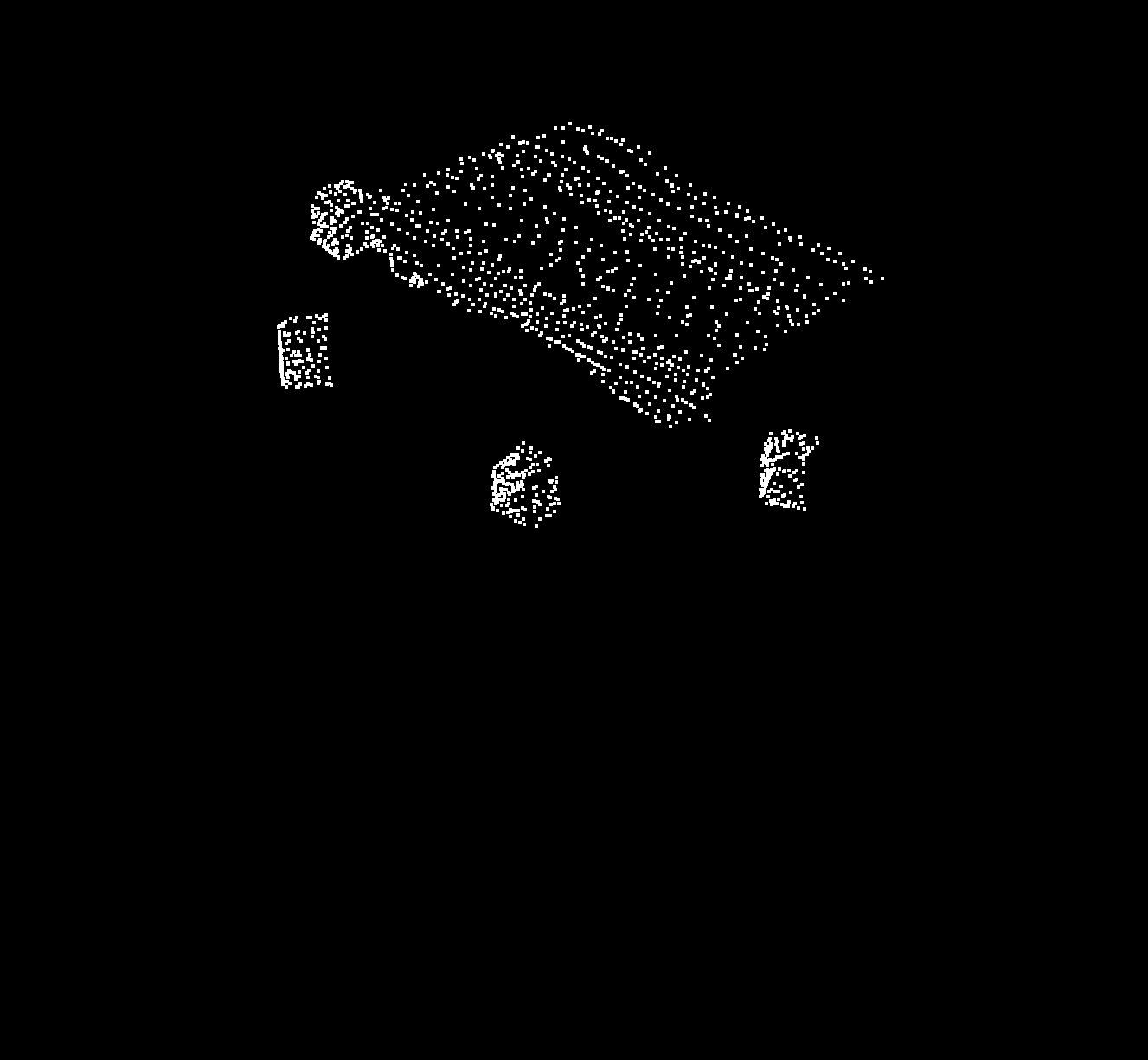}};

        \node[right=\xgap of m1] (m2)
            {\roundedimage[\imgw]
            {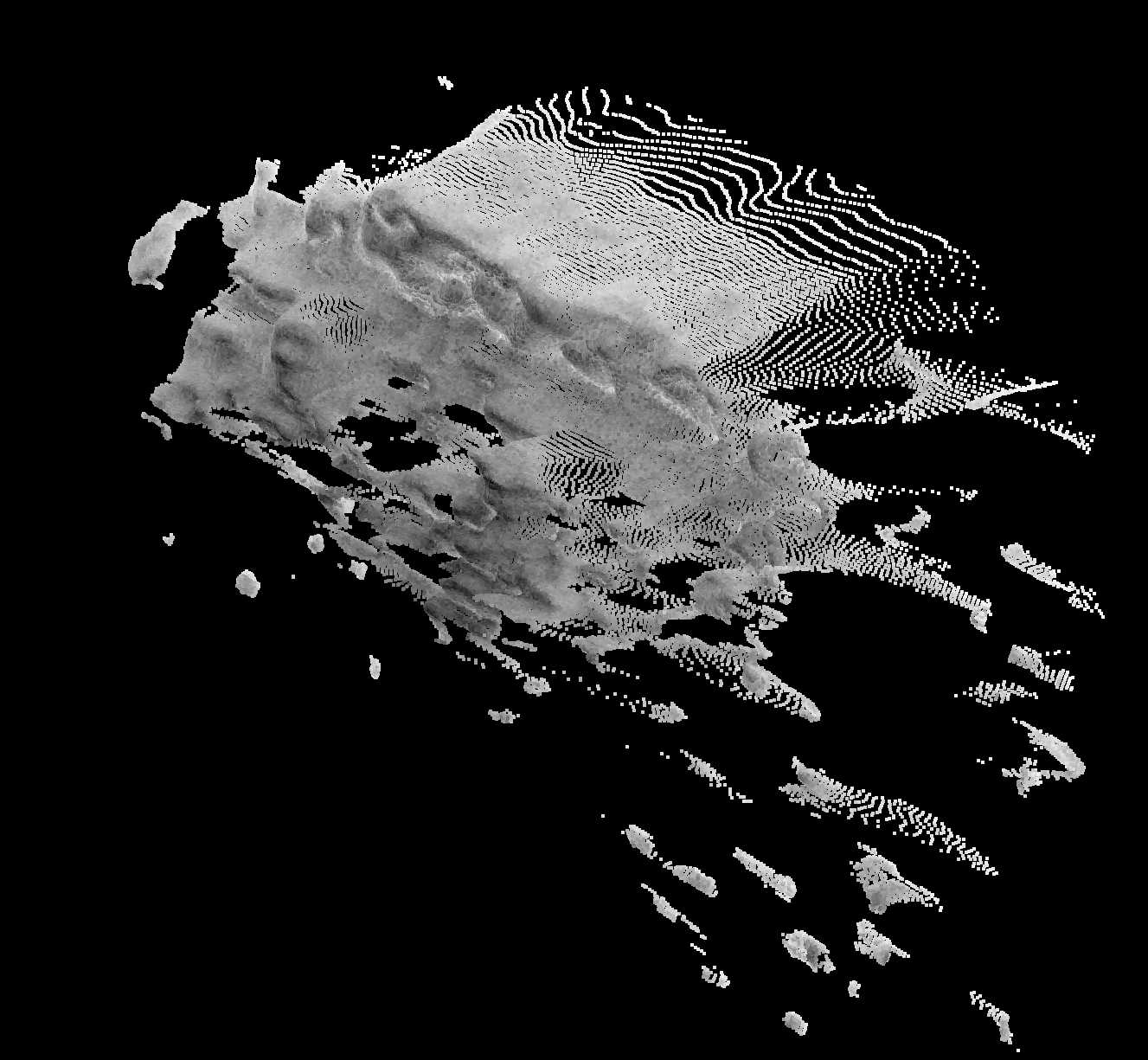}};

        \node[right=\xgap of m2] (m3)
            {\roundedimage[\imgw]
            {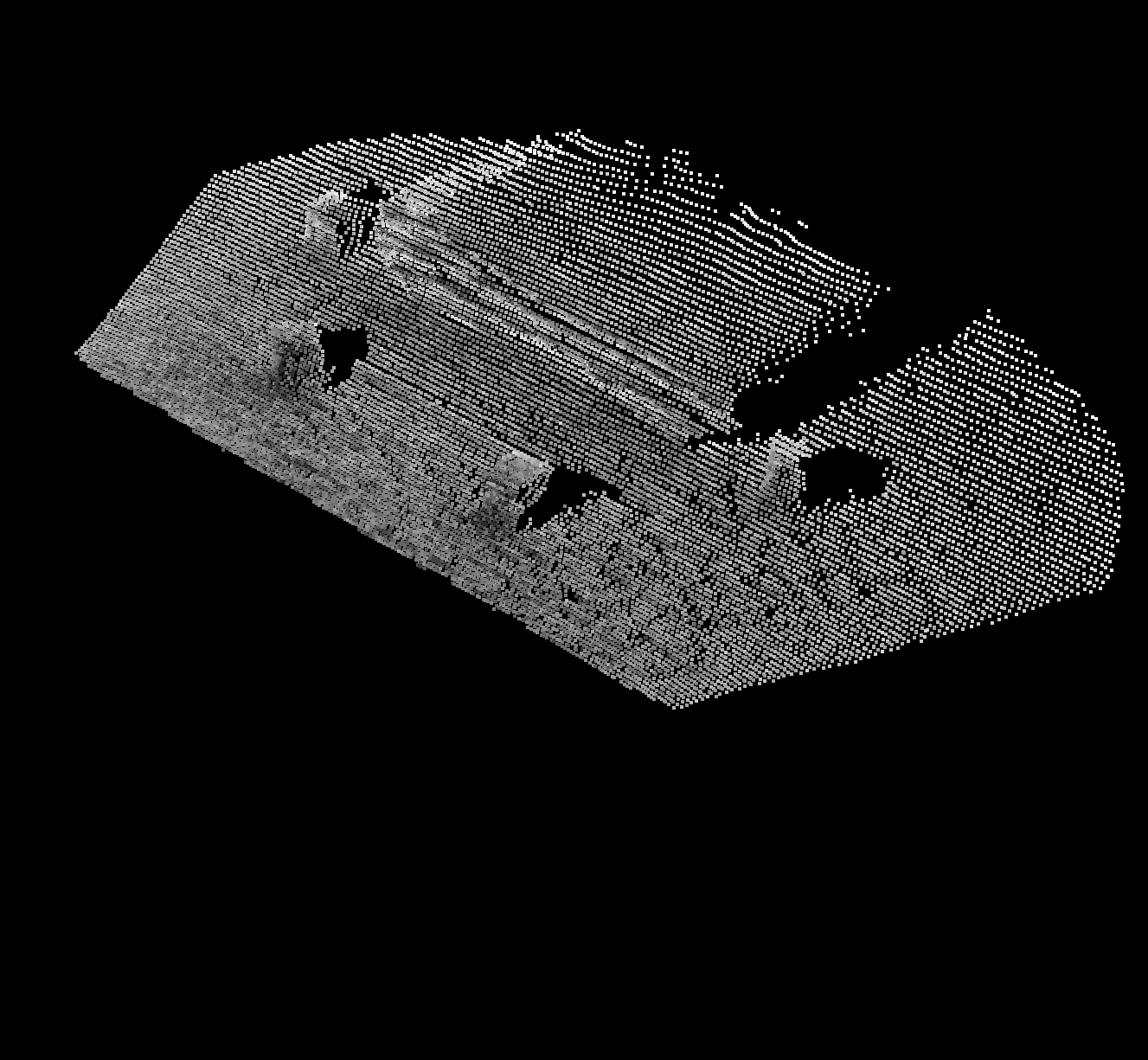}};

        \node[right=\xgap of m3] (m4)
            {\roundedimage[\imgw]
            {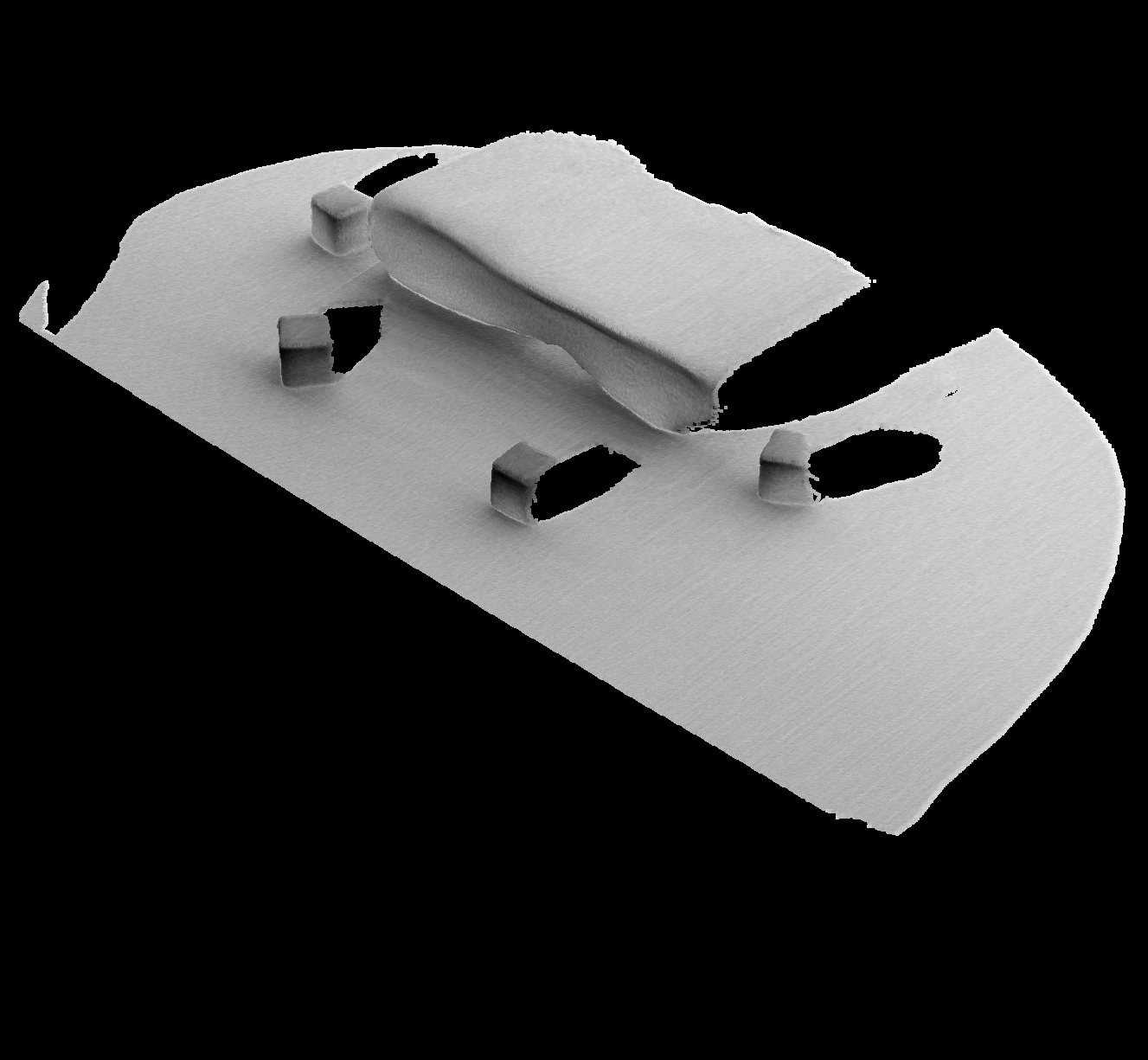}};

        \node[right=\xgap of m4] (m5)
            {\roundedimage[\imgw]
            {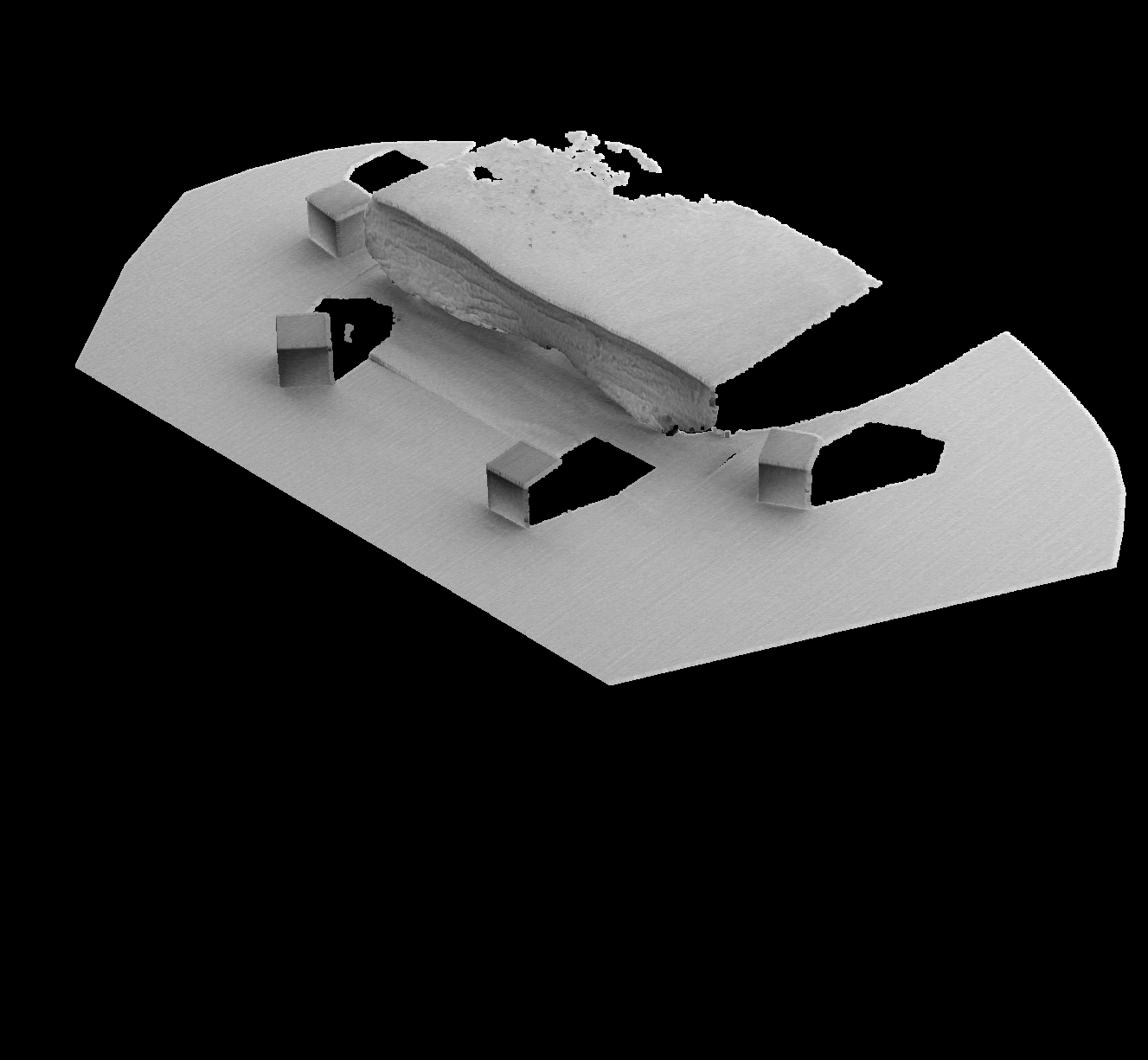}};

        \node[font=\bfseries, rotate=90]
            at ([xshift=-0.45cm]m1.west) {Porcine Belly};

        \node[below=\ygap of m1] (p1)
            {\roundedimage[\imgw]
            {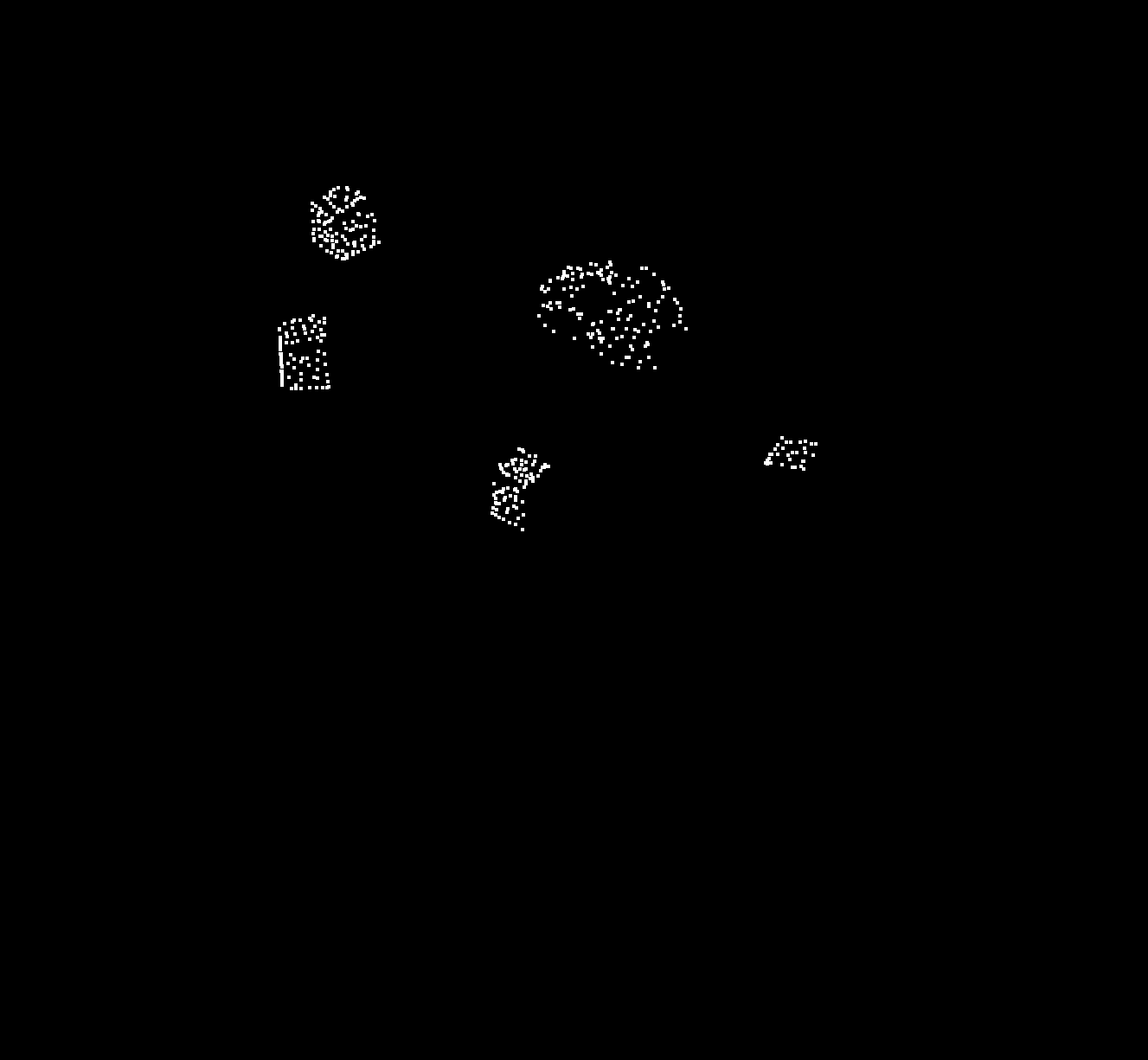}};

        \node[right=\xgap of p1] (p2)
            {\roundedimage[\imgw]
            {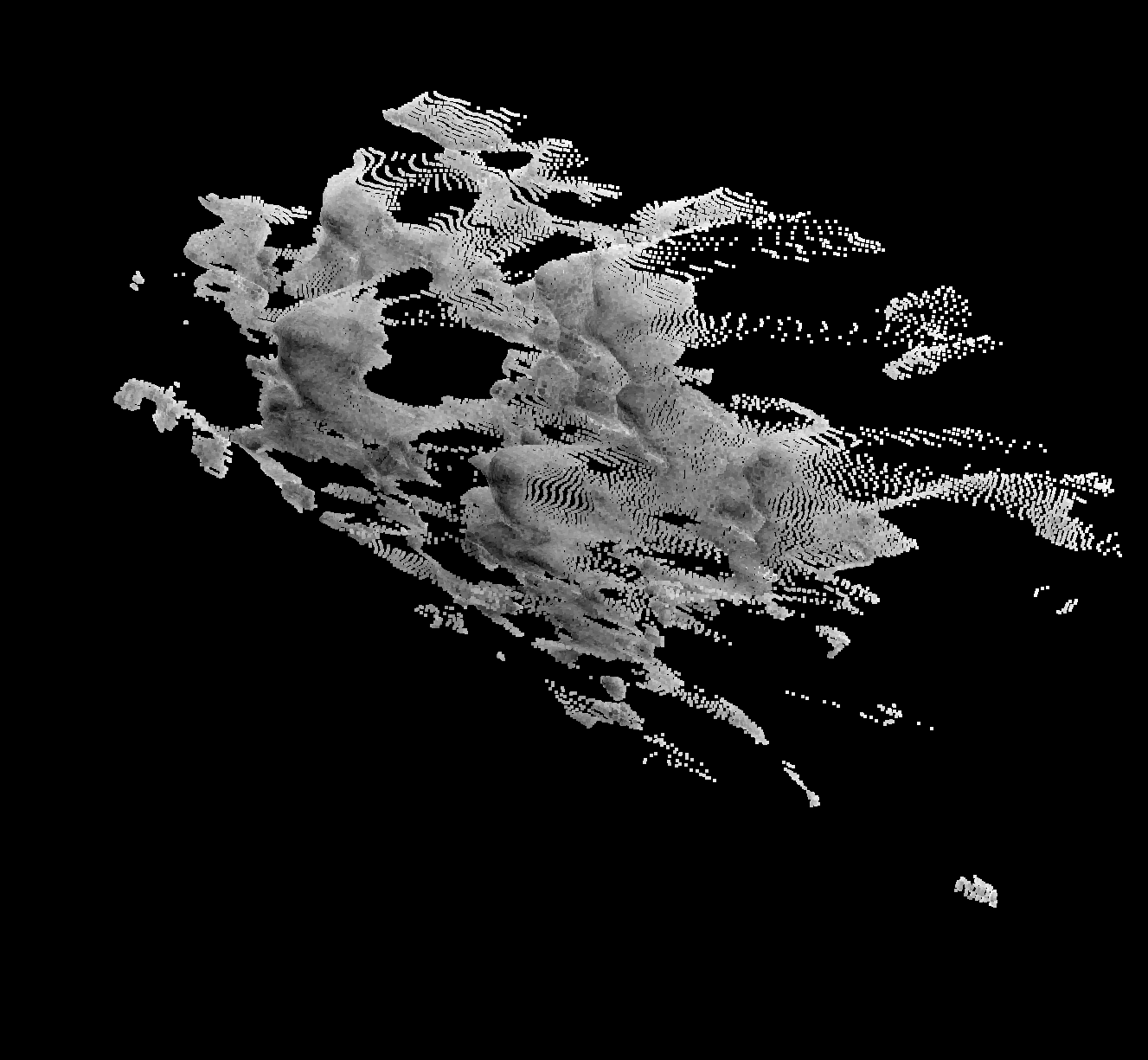}};

        \node[right=\xgap of p2] (p3)
            {\roundedimage[\imgw]
            {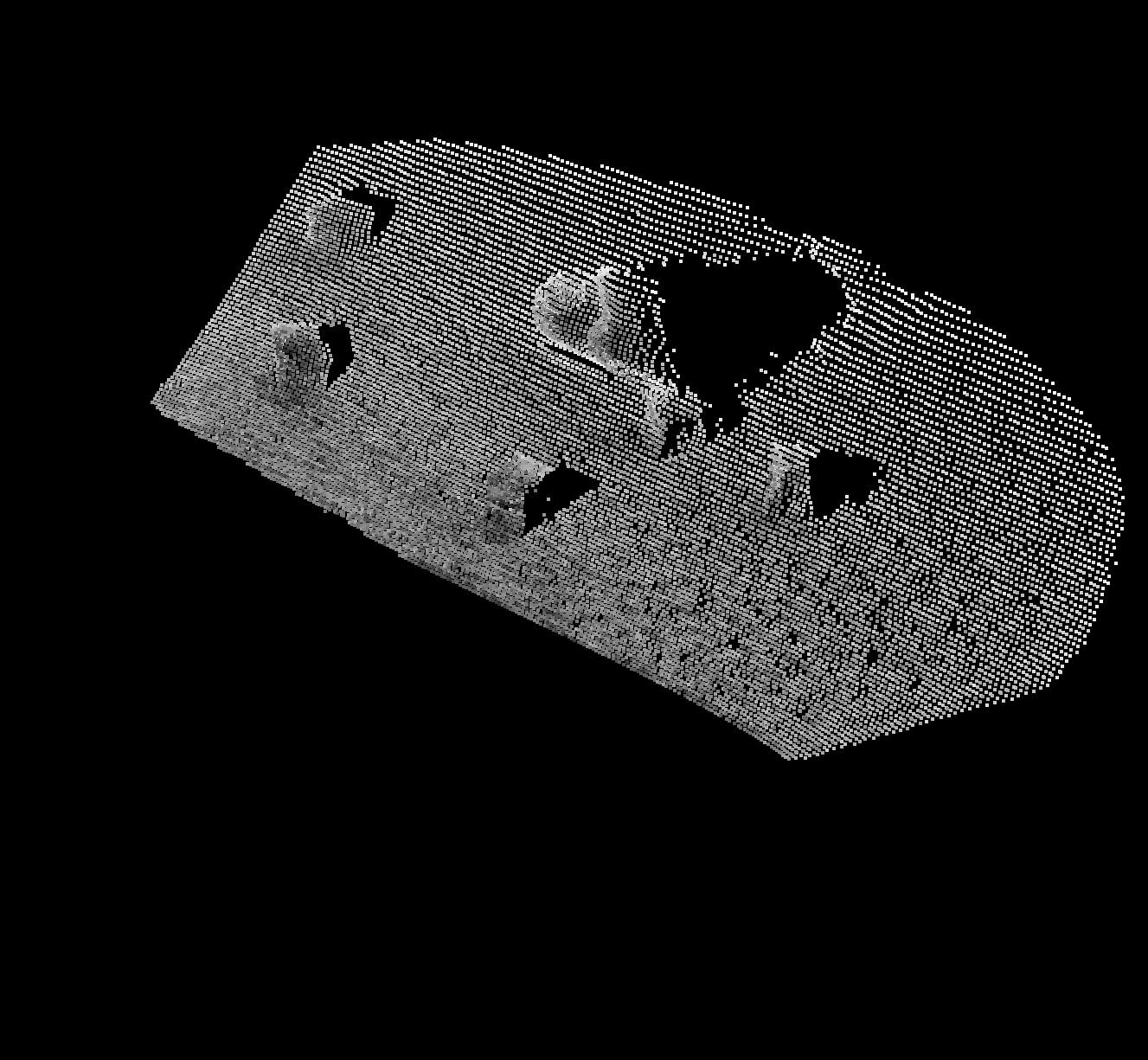}};

        \node[right=\xgap of p3] (p4)
            {\roundedimage[\imgw]
            {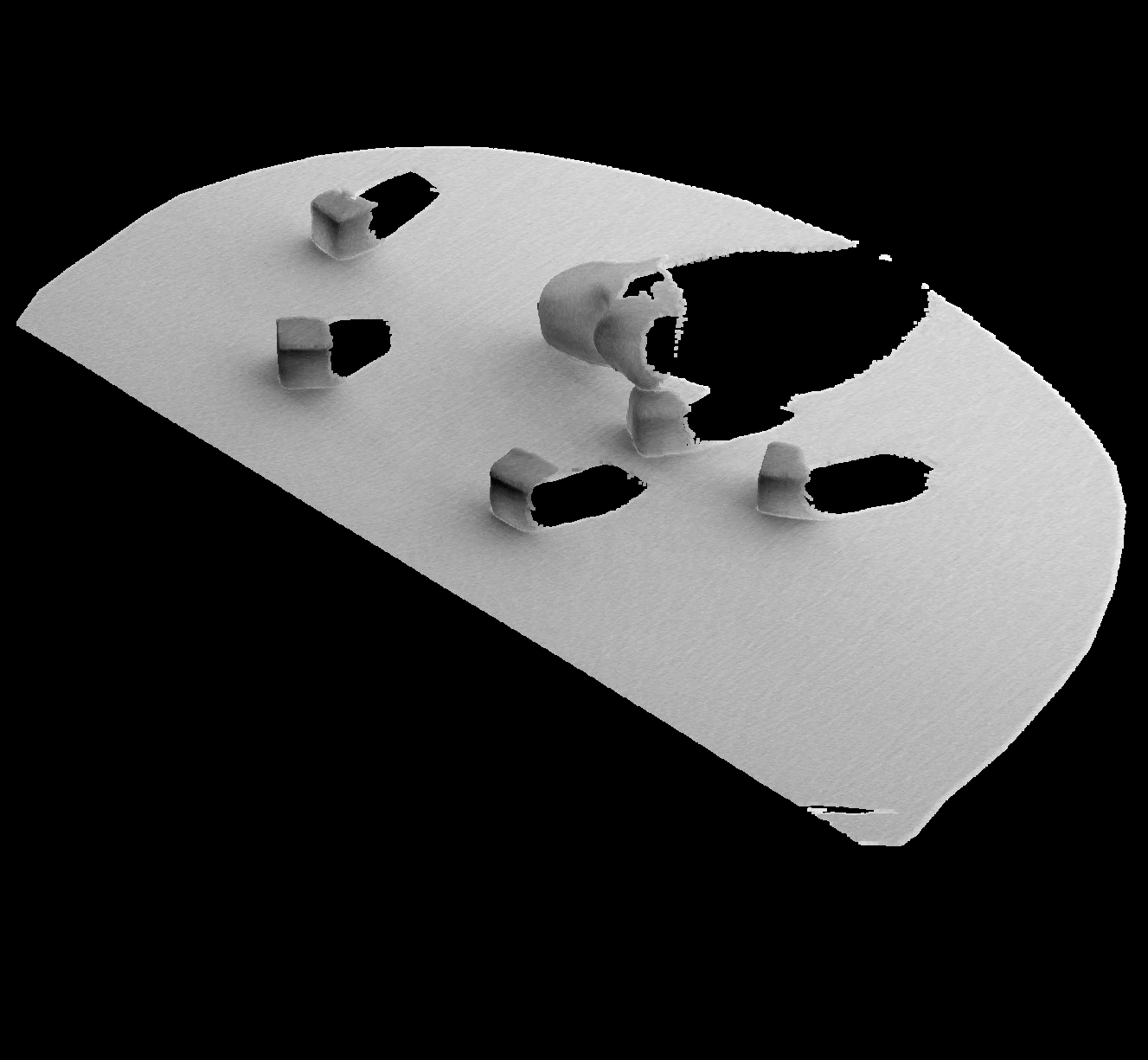}};

        \node[right=\xgap of p4] (p5)
            {\roundedimage[\imgw]
            {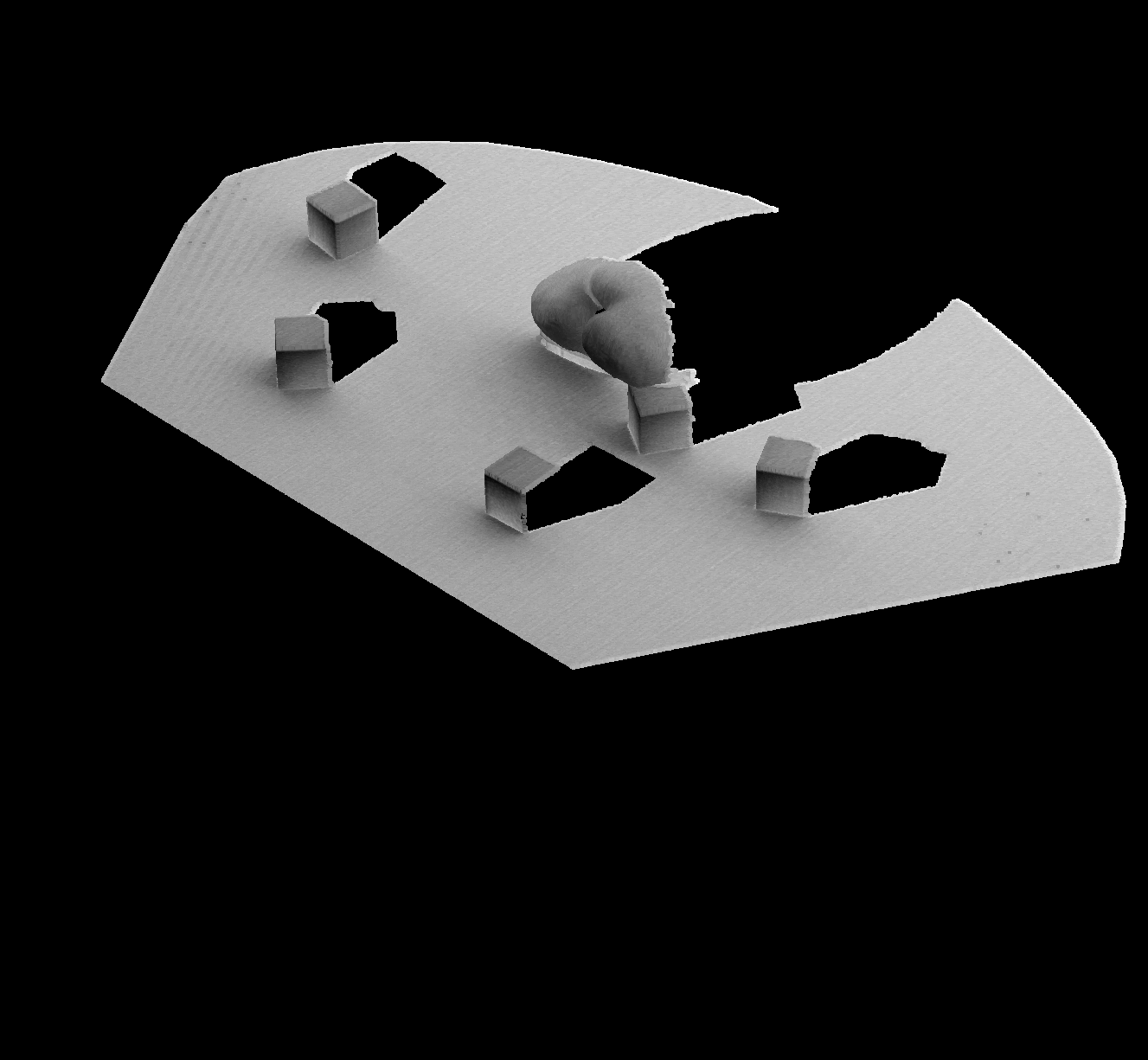}};

        \node[font=\bfseries, rotate=90]
            at ([xshift=-0.45cm]p1.west) {Phantom};

    \end{tikzpicture}

    \caption{
        Qualitative comparison of point clouds obtained from each depth sensor (structure only). All images were captured at approximately \SI{50}{\centi\metre} from the object. The backgrounds were trimmed away for visualization purposes.
    }

    \label{fig:all_point_clouds}

\end{figure*}

    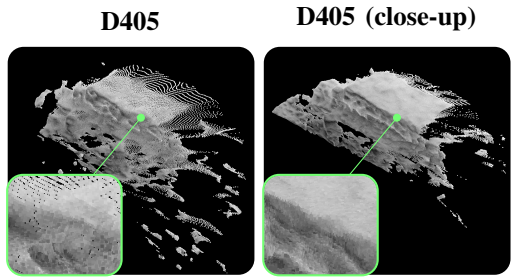
\begin{figure}[tb]
    \centering

    \begin{tikzpicture}[
        every node/.style={inner sep=0pt}
    ]

        \def\imgw{0.18\textwidth}

        \node (a)
            {\roundedimage[\imgw]
            {graphics/all_point_clouds/meat/d405.png}};

        \node[right=0.01\columnwidth of a] (b)
            {\roundedimage[\imgw]
            {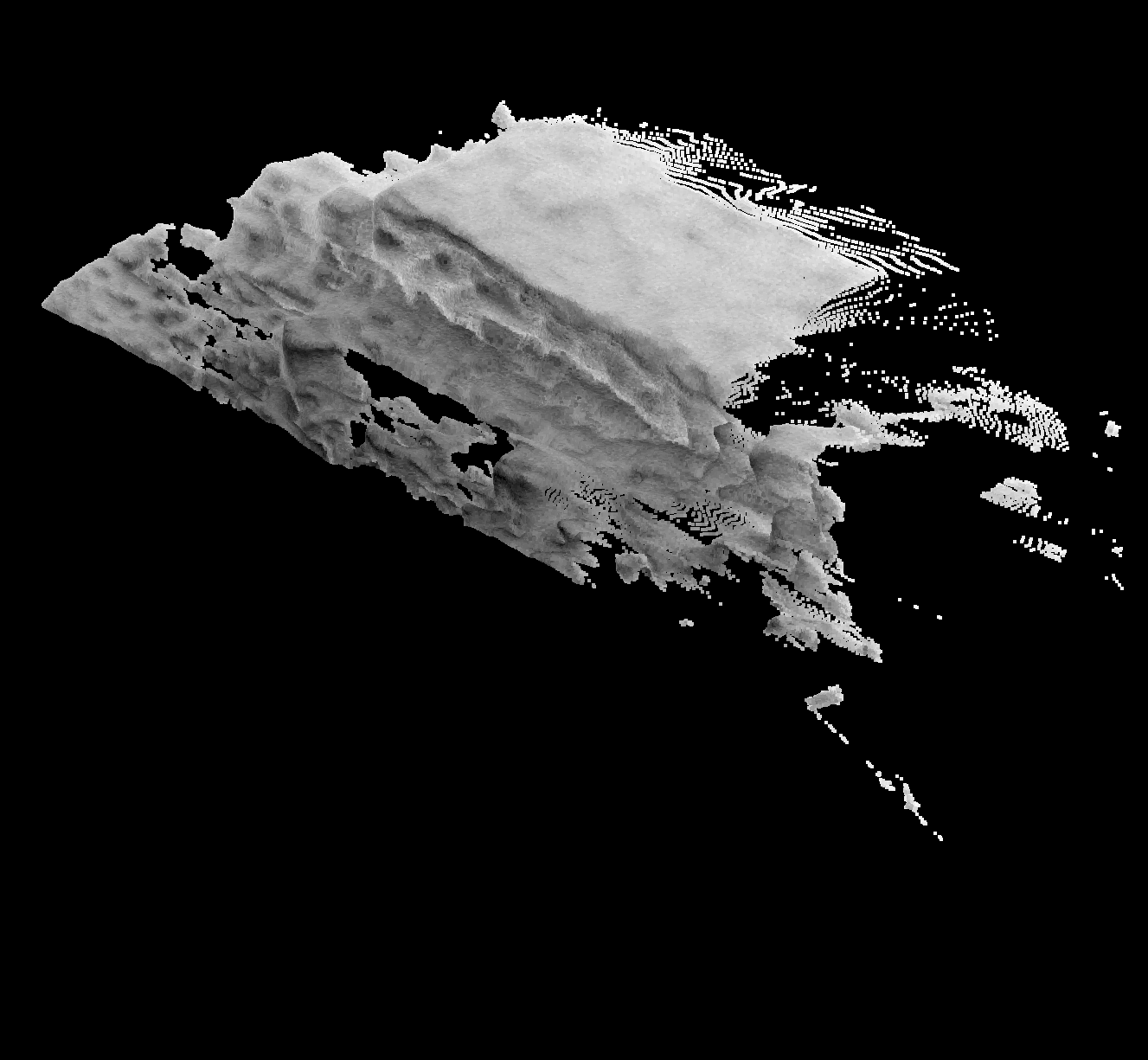}};

        \node[font=\bfseries, above=0.2cm of a] {D405};
        \node[font=\bfseries, above=0.2cm of b] {D405 (close-up)};

    \end{tikzpicture}

    \caption{
        Qualitative comparison of the \SI{50}{\centi\metre} and the \SI{33}{\centi\metre} close-up D405 point clouds (structure only) on the porcine belly. 
        The backgrounds were trimmed away for visualization purposes.
    }

    \label{fig:d405_closeup}

\end{figure}

\section{Results}

        \begin{table*}[tb]
    \centering
    \small
    \setlength{\tabcolsep}{3pt}
    \caption{Statistics over per-point errors for each camera and object. 
    Metrics are computed over three captures and averaged. \textbf{Kept} denotes the percentage of reference points that was retained after visibility filtering. Metrics are in \SI{}{\milli\metre} if not stated otherwise. 
    The last three columns are the number of points (in percentage) for which the per-point error is less than $2$, $5$, or \SI{10}{\mm}.}
    \label{tab:tukey-filtered-camera-metrics}
        \begin{tabular}{p{3cm}lrrrrrrrr}
        Object & Camera & Kept [\%] & Mean & Std. & Median & \SI{<2}{\mm} [\%] & \SI{<5}{\mm} [\%] & \SI{<10}{\mm} [\%] \\
        \hline
        Bone    & D405           & $ 99.3 $ & $ 7.11 $ & $ 3.52 $ & $ 6.97 $ & $  7.8 $ & $ 29.7 $ & $ 77.1 $ \\
        Bone    & Flexx          & $ 93.2 $ & $ 6.97 $ & $ 2.47 $ & $ 6.92 $ & $  1.3 $ & $ 22.1 $ & $ 87.3 $ \\
        Bone    & ZED            & $ 97.9 $ & $ 1.58 $ & $ 0.56 $ & $ 1.56 $ & $ 77.1 $ & $100.0 $ & $100.0 $ \\
        Bone    & Zivid          & $ 96.1 $ &  \textbf{1.20}  & \textbf{0.38}  &  \textbf{1.23}  & $ 97.8 $ & $100.0 $ & $100.0 $ \\
        \hline
        Porcine Belly    & D405           & $ 97.0 $ & $ 5.45 $ & $ 2.55 $ & $ 5.52 $ & $ 10.7 $ & $ 42.7 $ & $ 95.8 $ \\
        Porcine Belly    & D405 (close-up) & $ 99.2 $ & $ 3.28 $ & $ 1.69 $ & $ 3.06 $ & $ 29.3 $ & $ 82.1 $ & $100.0 $ \\
        Porcine Belly    & Flexx          & $ 93.4 $ & $ 6.98 $ & $ 2.15 $ & $ 6.93 $ & $  0.5 $ & $ 17.3 $ & $ 91.0 $ \\
        Porcine Belly    & ZED            & $ 91.7 $ & $ 1.60 $ & $ 0.51 $ & $ 1.59 $ & $ 79.4 $ & $100.0 $ & $100.0 $ \\
        Porcine Belly    & Zivid          & $ 91.5 $ &  \textbf{1.42}  &  \textbf{0.44}  &  \textbf{1.41}  & $ 90.5 $ & $100.0 $ & $100.0 $ \\
        \hline
        Phantom & D405           & $ 95.5 $ & $ 4.45 $ & $ 2.60 $ & $ 4.03 $ & $ 19.0 $ & $ 65.6 $ & $ 96.6 $ \\
        Phantom & Flexx          & $ 94.8 $ & $ 4.68 $ & $ 2.23 $ & $ 4.54 $ & $ 10.8 $ & $ 57.0 $ & $ 98.2 $ \\
        Phantom & ZED            & $100.0 $ & $ 5.99 $ & $ 2.64 $ & $ 5.62 $ & $  3.2 $ & $ 40.8 $ & $ 93.3 $ \\
        Phantom & Zivid          & $ 92.5 $ &  \textbf{2.21}  &  \textbf{0.83}  &  \textbf{2.17}  & $ 40.1 $ & $ 99.7 $ & $100.0 $ \\
        \end{tabular}
    \end{table*}

    The first point cloud captured by each sensor is shown in \Cref{fig:all_point_clouds}.
    \Cref{tab:tukey-filtered-camera-metrics} summarizes the quantitative results.
    Zivid achieved the lowest mean error on all objects, with \SI{1.2}{\milli\metre} on bone, \SI{1.42}{\milli\metre} on porcine belly, and \SI{2.21}{\milli\metre} on the phantom.
    ZED was close to Zivid on bone and porcine belly with mean errors of \SI{1.58}{\milli\metre} and \SI{1.60}{\milli\metre}, but it degraded to \SI{5.99}{\milli\metre} on the phantom.
    The D405 produced a mean error of \SI{5.45}{\milli\metre} on the porcine belly at \SI{50}{\centi\metre} and an error of \SI{3.28}{\milli\metre} at \SI{33}{\centi\metre}.
    Flexx performed similarly to the D405, with errors between $4.68$ and \SI{6.98}{\milli\metre}.
    For all objects and cameras, more than 90\% of reference points remained after visibility filtering.%

\section{Discussion and Conclusion}

    The results indicate that Zivid was the most consistent sensor in our experiments, while ZED was similarly accurate on bone and porcine belly but less reliable on the dark homogeneous phantom (with an error higher than all other sensors).
    The results show that working distance strongly affects the output of the D405; a closer porcine belly acquisition improves all error metrics noticeably and results in a clearer overall point cloud, see \Cref{fig:d405_closeup}.
    Still, the D405 point clouds shown in \Cref{fig:all_point_clouds} are noisy and contain many artifacts that are likely caused by the homogeneous white surface of the table which provides few features for traditional stereo systems.
    Our findings are contrary to the results from previous works~\cite{burger2023comparative, villa2025benchmarking} that recommend the Intel RealSense D405.
    We attribute this large difference to the distance at which we compare the sensors.
    The Flexx point clouds look well structured, but it performs noticeably worse than the ZED and the Zivid.
    Additionally, Flexx provides poor coverage of the porcine belly's rear part.
    Sparse areas in captured point clouds can artificially increase the reported errors as they are based on shortest distances computed from closest points.
    However, we found that a more aggressive pruning (decreasing the visibility threshold to $Q_3 + 0.1\cdot\mathrm{IQR}$) for porcine belly only reduced the mean error from \SI{6.98}{\milli\metre} to only \SI{6.20}{\milli\metre} while removing $25.9\%$ of points, implying that this is not the primary source of the high error.
    While the Zivid provides the overall best results, the used acquisition mode requires over one second to capture a single depth estimation and results in holes on the porcine belly, see \Cref{fig:all_point_clouds}.
    Moreover, and despite the Zivid having faster modes, the bright projected pattern may make the system unsuitable for some medical applications.
    
    Main limitations are the sparse reference geometries, the isolated-object scene design, the single viewpoint, and the use of single-sided error metrics, which evaluate accuracy on captured regions but do not penalize missing coverage.
    The conclusions apply to the evaluated device and do not establish general rankings of their underlying sensing principles.
    We also highlight that the errors of the Zivid and ZED cameras are close to the self-reported stylus tracking error.
    
    Future work should add bidirectional metrics to measure completion, additional viewpoints, realistic surgical clutter, and neural stereo reconstruction methods~\cite{wen2025foundationstereo}.

    Overall, this study compares commercial depth sensors on real animal tissue and shows that, at an operating distance of \SI{\sim50}{\centi\metre}, the Stereolabs ZED 2i and Zivid 2M+ 60 provide the best accuracy on the evaluated real tissue, while the Zivid should be preferred for homogeneous silicone phantoms.

\vspace {1cm}
\noindent\textsf{\textbf{Author Statement}}\\
This work is supported by the state of Bavaria through the Bayerische Forschungsstiftung (BFS) under Grant AZ-1592-23-ForNeRo.

\bibliographystyle{ieeetr}
\bibliography{STI}

@ARTICLE{henrich2024looc,
  author={Henrich, Pit and Mathis-Ullrich, Franziska},
  journal={IEEE Access}, 
  title={LOOC: Localizing Organs Using Occupancy Networks and Body Surface Depth Images}, 
  year={2025},
  volume={13},
  number={},
  pages={36930-36938},
  doi={10.1109/ACCESS.2025.3543736}}

@ARTICLE{henrich2025ludo,
  author={Henrich, Pit and Mathis-Ullrich, Franziska and Scheikl, Paul Maria},
  journal={IEEE Transactions on Robotics}, 
  title={LUDO: Low-Latency Understanding of Deformable Objects Using Point Cloud Occupancy Functions}, 
  year={2025},
  volume={41},
  number={},
  pages={4283-4299},
  doi={10.1109/TRO.2025.3582837}}

@inproceedings{wen2025foundationstereo,
  title={Foundationstereo: Zero-shot stereo matching},
  author={Wen, Bowen and Trepte, Matthew and Aribido, Joseph and Kautz, Jan and Gallo, Orazio and Birchfield, Stan},
  booktitle={Proceedings of the Computer Vision and Pattern Recognition Conference},
  pages={5249--5260},
  year={2025}
}

@ARTICLE{gyenes2025,
  author={Gyenes, Balázs and Franke, Nikolai and Scheikl, Paul Maria and Henrich, Pit and Younis, Rayan and Neumann, Gerhard and Wagner, Martin and Mathis-Ullrich, Franziska},
  journal={IEEE Robotics and Automation Letters}, 
  title={Point Cloud Segmentation for Autonomous Clip Positioning in Laparoscopic Cholecystectomy on a Phantom}, 
  year={2025},
  volume={10},
  number={8},
  pages={8522-8529},
  doi={10.1109/LRA.2025.3585357}}

@article{burger2023comparative,
  title={Comparative evaluation of three commercially available markerless depth sensors for close-range use in surgical simulation},
  author={Burger, Lukas and Sharan, Lalith and Karl, Roger and Wang, Christina and Karck, Matthias and De Simone, Raffaele and Wolf, Ivo and Romano, Gabriele and Engelhardt, Sandy},
  journal={International journal of computer assisted radiology and surgery},
  volume={18},
  number={6},
  pages={1109--1118},
  year={2023},
  publisher={Springer}
}

@article{villa2025benchmarking,
  title={Benchmarking commercial depth sensors for intraoperative markerless registration in neurosurgery applications},
  author={Villa, Manuel and Sancho, Jaime and Rosa-Olmeda, Gonzalo and Chavarrias, Miguel and Juarez, Eduardo and Sanz, Cesar},
  journal={International Journal of Computer Assisted Radiology and Surgery},
  volume={20},
  number={8},
  pages={1759--1769},
  year={2025},
  publisher={Springer}
}

@article{curto2022experimental,
  title={An experimental assessment of depth estimation in transparent and translucent scenes for Intel RealSense D415, SR305 and L515},
  author={Curto, Eva and Araujo, Helder},
  journal={Sensors},
  volume={22},
  number={19},
  pages={7378},
  year={2022},
  publisher={MDPI}
}

@article{servi2021metrological,
  title={Metrological characterization and comparison of d415, d455, l515 realsense devices in the close range},
  author={Servi, Michaela and Mussi, Elisa and Profili, Andrea and Furferi, Rocco and Volpe, Yary and Governi, Lapo and Buonamici, Francesco},
  journal={Sensors},
  volume={21},
  number={22},
  pages={7770},
  year={2021},
  publisher={MDPI}
}

@inproceedings{max1,
  title={IRBSM: a Deep implicit 3D breast shape model},
  author={Weiherer, Maximilian and von Riedheim, Antonia and Br{\'e}bant, Vanessa and Egger, Bernhard and Palm, Christoph},
  booktitle={BVM Workshop},
  pages={38--43},
  year={2025},
  organization={Springer}
}

\end{document}